\documentclass[runningheads]{llncs}

\usepackage[mobile]{eccv}
\usepackage{eccvabbrv}

\usepackage{graphicx}
\usepackage{booktabs}

\usepackage[accsupp]{axessibility}  % Improves PDF readability for those with disabilities.

\usepackage[pagebackref,breaklinks,colorlinks,citecolor=eccvblue]{hyperref}
\usepackage{orcidlink}

\usepackage{xcolor}

\usepackage{amsmath,amsfonts,bm}

\def\eqref#1{equation~\ref{#1}}
\def\1{\bm{1}}

\DeclareMathAlphabet{\mathsfit}{\encodingdefault}{\sfdefault}{m}{sl}
\SetMathAlphabet{\mathsfit}{bold}{\encodingdefault}{\sfdefault}{bx}{n}

\usepackage{graphicx}
\usepackage{tabularx}
\usepackage{adjustbox}
\usepackage{subcaption}
\usepackage{array}
\usepackage{multirow}
\usepackage{colortbl}
\usepackage{wrapfig}
\usepackage{booktabs}
\usepackage{enumitem}
\usepackage{csquotes}
\usepackage{tikz}
\usepackage{pdflscape}
\usepackage{natbib}
\usetikzlibrary{shapes.geometric}
\usetikzlibrary{fit}
\usetikzlibrary{positioning}

\definecolor{watercolor}{HTML}{0000FF}
\definecolor{TropicalSubtropicalMoistBroadleafForests}{HTML}{38A700}
\definecolor{TropicalSubtropicalDryBroadleafForests}{HTML}{CCCD65}
\definecolor{TropicalSubtropicalConiferousForests}{HTML}{88CE66}
\definecolor{TemperateBroadleafMixedForests}{HTML}{00734C}
\definecolor{TemperateConiferForests}{HTML}{458970}
\definecolor{BorealForestsTaiga}{HTML}{7AB6F5}
\definecolor{TropicalSubtropicalGrasslandsSavannasShrublands}{HTML}{FEAA01}
\definecolor{TemperateGrasslandsSavannasShrublands}{HTML}{FEFF73}
\definecolor{FloodedGrasslandsSavannas}{HTML}{BEE7FF}
\definecolor{MontaneGrasslandsShrublands}{HTML}{D6C39D}
\definecolor{Tundra}{HTML}{9ED7C2}
\definecolor{MediterraneanForestsWoodlandsScrub}{HTML}{FE0000}
\definecolor{DesertsXericShrublands}{HTML}{CC6767}
\definecolor{Mangroves}{HTML}{FE01C4}

\begin{document}

\title{SatCLIP: Global, General-Purpose Location Embeddings with Satellite Imagery} 

\titlerunning{SatCLIP}

\author{Konstantin Klemmer\inst{1}\orcidlink{0000-0002-7096-0133} \and
Esther Rolf\inst{2,3}\orcidlink{0000-0001-5066-8656} \and
Caleb Robinson\inst{4}\orcidlink{0000-0003-1975-4454} \and
Lester Mackey\inst{1}\orcidlink{0000-0002-1102-0387} \and
Marc Rußwurm\inst{5}\orcidlink{0000-0001-6612-5744}} 

\authorrunning{K.~Klemmer et al.}

\institute{Microsoft Research New England, Cambridge, MA, USA \and University of Colorado Boulder, CO, USA \and
Harvard University, Cambridge, MA, USA \and
Microsoft AI for Good Research Lab, Redmond, WA, USA \and
Wageningen University, Wageningen, Netherlands
}

\maketitle
\begin{abstract}
Geographic information is essential for modeling tasks in fields ranging from ecology to epidemiology. However, extracting relevant location characteristics for a given task can be challenging, often requiring expensive data fusion or distillation from massive global imagery datasets. To address this challenge, we introduce Satellite Contrastive Location-Image Pretraining (SatCLIP). This \emph{global, general-purpose geographic location encoder} learns an implicit representation of locations by matching CNN and ViT inferred visual patterns of openly available satellite imagery with their geographic coordinates. The resulting SatCLIP location encoder efficiently summarizes the characteristics of any given location for convenient use in downstream tasks. In our experiments, we use  SatCLIP embeddings to improve prediction performance on nine diverse location-dependent tasks including temperature prediction, animal recognition, and population density estimation.  
Across tasks, SatCLIP consistently outperforms alternative location encoders and improves geographic generalization by encoding visual similarities of spatially distant environments.  These results demonstrate the potential of vision-location models to learn meaningful representations of our planet from the vast, varied, and largely untapped modalities of geospatial data.
%including models trained on natural images and those trained on text. 
%(e.g., the same climate zone on different continents)
\end{abstract}

\section{Introduction}
\label{sec1}
\begin{figure}[t]
    \centering
    \includegraphics[width=.9\textwidth]{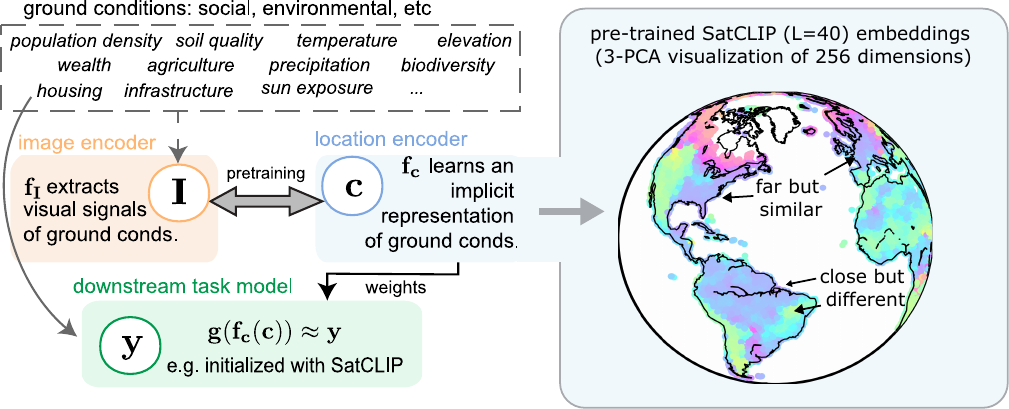}
    \caption{\textbf{Motivation for SatCLIP:} Capturing ground conditions from satellite images and transferring them into a location encoder via constrastive image-location pretaining. The right globe shows a PCA representation of the pre-trained location encoder.}
    \label{fig:satclip_intuition_figure}
\end{figure}

%Much of the world's data is geospatial and simultaneously captured by images. For instance, cellphone images can be used to capture the trajectories of taxis.
Satellite imagery has proven to be a valuable source of input data for predictive models across a wide range of real-world applications \citep{Rolf2021}, for example, interpolating missing air pollution data \citep{Chen2019}, crop yield forecasting \citep{Tseng2022,lobell2015scalable}, and agro-forestry carbon stock prediction \citep{Reiersen2022}. Many geospatial modeling tasks, in fields ranging from epidemiology, the Earth system sciences, to ecology also directly leverage geographic location for improving predictions \citep{Aodha2019, cole2023spatial}.
%In development settings, satellite-based machine learning models have helped to support humanitarian aid efforts by identifying dwellings in deprived areas \citep{Tiede2021,Wurm2019,Klemmer2020a}, predicting poverty \citep{Jean2016}, and mapping rural populations \citep{Hu2019}. 
%The location at which satellite images are taken maps them in the same geometric space: planet Earth. 
%
Patterns extracted from satellite images can describe the unique characteristics of locations, by capturing their natural and built environment. These characteristics are often correlated in space: While two nearby locations are more likely to have similar features (e.g. the same land cover), two distant locations can also share location characteristics when they share similar environmental ground conditions like climate zones (\cref{fig:satclip_intuition_figure}, right). %On the other hand, two distant locations that fall in, e.g., different climate zones, usually have very different visual appearances (e.g., a tropical rainforest vs an arctic tundra). 
Since the spatial patterns governing different geographic data modalities are often complex and non-linear, predictive models working with geo-data benefit from explicitly integrating intuitions for spatial and spatio-temporal dependencies \citep{fotheringham2009geographically, alam2022survey, Klemmer2021a, Klemmer2022, cole2023spatial}.

But integrating geographic information into a deep learning model is not straightforward. Even though spatial coordinates are often informative, introducing them as features can amplify geographic distribution shift problems and lead to poor evaluation accuracy. This is especially of concern for cross-regional generalization when data from evaluation areas (and their coordinates) are absent in the training data. Consequently, many location-informed models are only applicable for interpolation problems where the evaluation area overlaps with the training area. And while some settings warrant interpolation methods---e.g., species distribution modeling \citep{Aodha2019,cole2023spatial,berg2014birdsnap}, where data is global and somewhat representative spatially---for many applications, available labeled data is patchy and sparse, and predictive models must generalize to unseen geographic areas potentially far from the training data.
In these situations, utilizing location embeddings that capture ground conditions might be preferable to using raw spatial coordinates. We tackle this research problem by pretraining location encoders with globally and uniformly sampled satellite imagery and with general-purpose use in mind. Specifically, we leverage the location information indexing satellite images as an input to a contrastive pretraining objective that aims to match location-image pairs. This is analogous to the vision-language pretraining deployed in Contrastive Location Image Pretraining (CLIP) \citep{Radford2021}. %Satellite imagery captures location characteristics (e.g., vegetation and built structures) and has proven informative in diverse downstream tasks ranging from air pollution modeling \citep{Chen2019} to agricultural monitoring \citep{Tseng2022}. 

% commented out, this paragraph is similar to relatd work paragraph below
%Prior investigations of pretrained location encoders have been task-specific, but hint at their potential for general-purpose use. \Citet{Mai2023} pretrain location encoders for specific applications: species distribution mapping and land cover classification. They don't investigate the usability of their location encoders for new, out-of-domain tasks or for global generalization. In contrast, \citet{Yin2019} introduce GPS2Vec with general applications in mind. However, their approach requires training separate location encoders for each UTM zone from scratch. This permits them to obtain only location encoders for areas in which they have pretraining data, preventing global-scale applications and generalization across UTM-zones.

% copied paragraph from related work
\paragraph{Previous work on pretrained location encoders.} Three studies have pretrained geographic location encoders that input spatial coordinates, and return learned contextual representations. \Citet{Yin2019} propose GPS2Vec, a set of UTM-zone specific location encoders using geotagged Flickr images (YFCC100M) \citep{Thomee2016} and their corresponding semantic tags for training. Geographic generalization was out of scope for this work as embeddings are only available for UTM-zones in which training data can be found. \Citet{Mai2023} introduce Contrastive Spatial Pre-Training (CSP) on the iNaturalist 2018 (iNat) \citep{Horn2018} species imagery and the Functional Map of the World (FMoW) \citep{Christie2018} satellite image datasets. CSP is used for unsupervised pretraining and downstream prediction on the same datasets and was not conceptualized for use on other tasks. The CSP pretraining datasets, iNaturalist and FMoW, are also unevenly distributed over space, with high image densities in North America and Europe and few images outside of Western regions. \Citet{Cepeda2023} propose GeoCLIP, in which the authors pretrain image and location encoders using Flickr images from the MediaEval Placing Tasks 2016 (MP-16) dataset \citep{Larson2017}--another dataset with strong overrepresentation of Western countries. GeoCLIP is developed for the task of geo-locating (natural) images and is not optimized or tested for general-purpose use.

\paragraph{Aim and contributions. } Existing work leaves two important gaps: understanding how location encoders generalize across various downstream tasks and ensuring global coverage and approximately equal performance of location embeddings. In this work, we address both of these challenges by introducing the first \emph{global-coverage, general-purpose pretrained geographic location encoder} based on Satellite Contrastive Location Image Pretraining -- \textbf{SatCLIP}.
%We overcome the shortcomings of existing pretrained location encoders by designing, building, and analyzing the--to our knowledge--first \emph{general-purpose pretrained geographic location encoder with global coverage}--\textbf{SatCLIP}. 
%
SatCLIP distills spatially varying visual patterns from globally-distributed satellite data into an implicit neural representation in a comparatively small and efficient neural network. This location encoder projects a latitude and longitude coordinate into a higher-dimensional vector representation that is matched with a visual vector representation from a computer vision encoder (CNN or ViT), as detailed later in \cref{sec3}. % shown in the \enquote{pretraining} panel in \cref{fig:lgm}. 
%The SatCLIP pretraining objective captures the spatial variability in the appearances of images and converts them into an implicit neural representation within the location encoder. 
%For downstream tasks (right panel), we use the pre-trained location encoder to convert raw location coordinates into vectors that can be informative for a wide range of environmental and socioeconomic prediction tasks. 
%Here, \cref{fig:satclip_intuition_subfigure} and demonstrated with experiments in \cref{sec4}. 
More generally, the proposed framework represents a step towards geographically-informed \enquote{foundation models} trained with large, unlabeled datasets, that are usable for a wide range of tasks, and extrapolate to unseen geographic areas. Our contributions can be summarized as follows:

\begin{itemize}[leftmargin=.5cm]
    %\item We outline the concept of \emph{pretrained geographic location encoders with global coverage}. These are general-purpose models that coalesce geographic data into meaningful location embeddings, instantiated as pretrained neural networks that take global locations (latitude / longitude coordinates) as input.
    \item We develop the first task-generalizable, global-coverage location encoder--\textbf{SatCLIP}--trained on Sentinel-2 multispectral satellite imagery. We release the pretrained encoder as a \texttt{PyTorch} model. We also release our new pretraining dataset, \textbf{S2-100K}. 
    \item We compare \textbf{SatCLIP} with existing pretrained location encoders and other geographic feature generation approaches on nine diverse downstream tasks, ranging from temperature prediction to population density estimation, highlighting superior performance in prediction and geographic generalization.
\end{itemize}

%\section{Related Work}
%\label{sec2}
%\input{related_work}

\section{Satellite Contrastive Location-Image Pretraining}
\label{sec3}
\newcommand{\Enc}{f}
\newcommand{\loc}{\textup{c}}
\newcommand{\cont}{\textup{I}}

\begin{figure*}[t]
    %\begin{subfigure}{0.31\linewidth}
    %    \centering
    %    \includegraphics[height=1.15in]{figures/satclip_intuition_v2.pdf}
    %    \caption{Intuition behind SatCLIP.}
    %    \label{fig:satclip_intuition_subfigure}
    %\end{subfigure}
     %\begin{subfigure}{0.8\linewidth}
     %   \centering
        \centering
        \includegraphics[width=\textwidth]{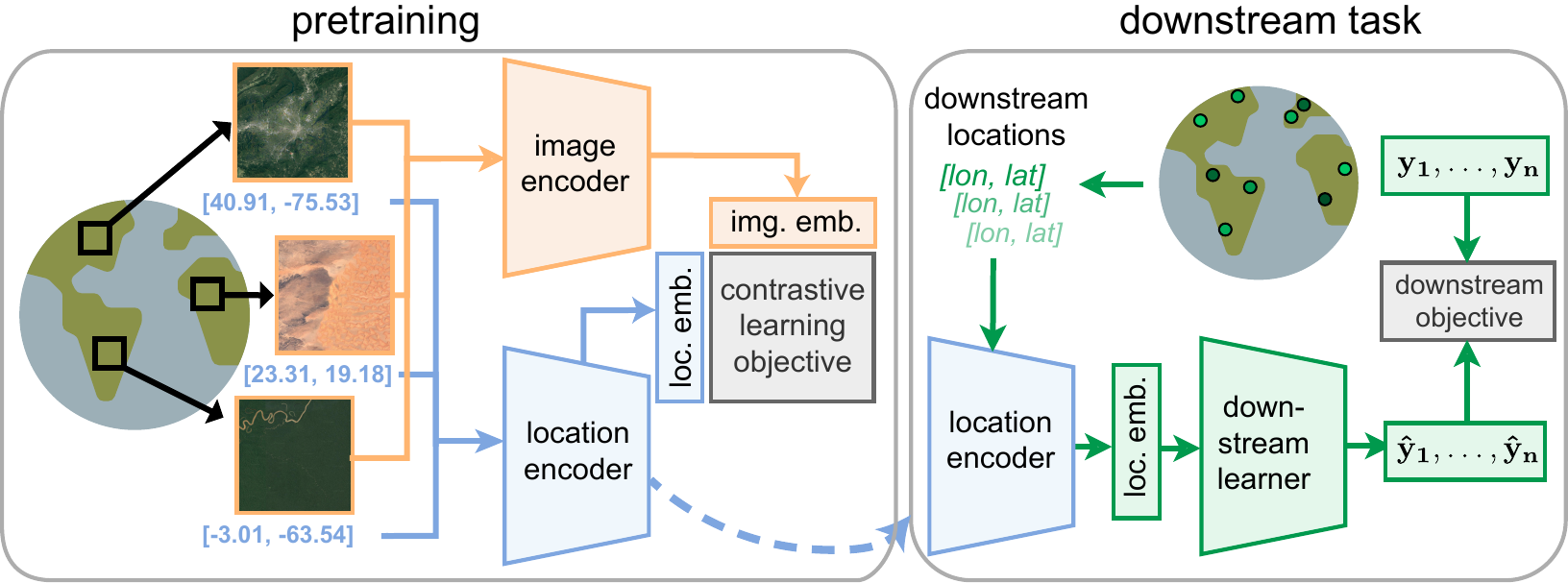}
    %    \caption{Global, general-purpose pretrained location encoders.}
        
    %\end{subfigure}
    \caption{\textbf{The SatCLIP pretraining and deployment pipeline.} 
    SatCLIP pretraining through image-location matching is outlined on the left. The pretrained location encoder can then be used in downstream tasks, highlighted on the right.}
    \label{fig:lgm}
    %\label{fig:SatCLIPIntuition}
\end{figure*}

With \textbf{SatCLIP}, we aim to train models that (1) provide \emph{general purpose embeddings} and (2) are \emph{globally representative}. In this section, we first motivate SatCLIP and then outline its components and training paradigm. %We illustrate the full SatCLIP pretraining and deployment pipeline in \cref{fig:lgm}. 

Various factors influence the appearance of Earth's surface, as captured in satellite images. In \cref{fig:satclip_intuition_figure}, we highlight that diverse environmental and socioeconomic factors are reflected in visual markers, such as the appearance of mountain ranges with their specific vegetation, the geometry and structure of agricultural fields, and the design of buildings. By training on the SatCLIP matching objective%(detailed later in \cref{eq:cliploss})
, the image encoder $\Enc_{\cont}$ is trained to associate an image $\mathbf{I}$ with a location $\mathbf{c}$ based on the various ground factors detectable in the images. At the same time, the location encoder $\Enc_{\loc}$ is trained to associate a given location with location-specific image characteristics. Effectively, both models learn to align the embeddings of a location and its corresponding image to maximize their similarity, as shown in the left panel of \cref{fig:lgm}. Inductive biases in the location encoder model control spatial smoothness (i.e. the $L$ hyperparameter in SatCLIP), allowing the model to interpolate to areas where no image-location pairs are present in the training data. After successful training, location features can be extracted at arbitrary locations. These features can be used to train any downstream learner $g$ that takes locations as input (right panel of \cref{fig:lgm}).

There are several advantages to training downstream models with location embeddings, as opposed to raw coordinates or images extracted at downstream locations. Models trained on raw coordinates $\mathbf{c}$ solely rely upon spatial dependencies without taking any ground conditions into account, such as local elevation patterns or local climate zones. Models trained on full images $\mathbf{I}$, while able to capture ground conditions, require expensive data preprocessing (downloading images for every downstream location) and training of large vision models. This can be infeasible in many, especially low-resource settings. SatCLIP has the potential to provide the best of both worlds: location embeddings capture both spatial effects and ground conditions while also being relatively low-dimensional (our SatCLIP location embeddings are $256$-dimensional vectors) and runtime efficient due to the small size of the location encoder model. This is particularly helpful in resource-constrained settings and allows fast encoding of coordinates to embedding space without GPUs.

%The pretrained representation models $\Enc_{\loc}$ can then be used to build spatially-aware downstream learners $g$. In our experiments, we instantiate $g$ by fine-tuning a model on top of $\Enc_{\loc}$ embeddings with annotated data in new downstream prediction tasks.

% \textcolor{red}{There are several advantages to representing locations with a direct model of the form $\Enc_{\loc}$, so long as that model can capture variation in key ground conditions across the globe. [ER: I think we need a paragraph like this that would discuss]
% \begin{itemize}
%     \item usability: (i) back of envelope runtime comparions and (ii)explicitly outline that people don't need imagery to use satclip downstream. 
%     \item explain that the important part is in how the location encoder embeds global data during pretraining (maybe allude to naive latlon strategies in the appendix)
% \end{itemize}
% }

\subsection{Pretraining with the SatCLIP Objective} %Location and Image Encoders}\label{sec:pretrained}

The inputs to a geographic location encoder are latitude/longitude coordinate pairs $\textbf{c}_{i} = [\lambda_{i}, \phi_{i}]$, where $\lambda_i$ is the longitude, $\phi_i$ is the latitude, and $i$ indexes locations on the spherical surface $\mathbb{S}^2$. For each location $i$, we have a corresponding multi-spectral image $\mathbf{I}_i \in \mathbb{R}^{m \times n \times k}$ with $k$ channels. We now define two encoders, a \emph{location encoder} $\Enc_{\loc}: \mathbb{S}^2 \rightarrow \mathbb{R}^{d}$ that takes in $2$-dimensional coordinates $\textbf{c}_{i}$ and returns a $d$-dimensional latent embedding and a \emph{image encoder} $\Enc_{\cont}: \mathbb{R}^{m{\times n \times k}} \rightarrow \mathbb{R}^{d}$ that takes in an image $\textbf{{I}}_{i}$ and also returns a $d$-dimensional latent embedding. 
%Note that the image encoder can be seen as a special case of a more general \textsl{context} encoder that may integrate other location-specific data modalities like audio or, in the case of GPS2Vec, text--or multiple modalities at once.

% \subsection{Pretraining with the SatCLIP Objective}

We train both encoders with the simple but highly effective CLIP \citep{Radford2021} objective
\begin{align}
\label{eq:cliploss}
\mathcal{L} \!= \!\frac{1}{2N}\!\left[\sum_{i=1}^N\!\mathcal{L}_\textup{loc}(\mathbf{c}_i, \mathbf{I}_{1,\dots,N})\! +\! \sum_{i=1}^N\!\mathcal{L}_\textup{img}(\mathbf{I}_i, \mathbf{c}_{1,\dots,N})\right]
\end{align}
that matches each coordinate $\mathbf{c}_i$ with the corresponding image $\mathbf{I}_i$ and against all images $\mathbf{I}_{1,\dots,N}$ using
\begin{align}
\textstyle
\mathcal{L}_\textup{loc}(\mathbf{c}_i, \mathbf{I}_{1,\dots,N}) = -\log \frac{\exp\left(\left\langle\Enc_{\loc}(\textbf{c}_i), \Enc_{\cont}(\textbf{I}_i)\right\rangle / \tau\right)}{\sum_{j=1}^{N} \exp\left(\left\langle\Enc_{\loc}(\textbf{c}_i), \Enc_{\cont}(\textbf{I}_{j})\right\rangle / \tau\right)}
\end{align}
and each image with the corresponding coordinate using a complementary loss $\mathcal{L}_\textup{img}(\mathbf{I}_i, \mathbf{c}_{1,\dots,N})$
%\begin{align}
%\textstyle
%\mathcal{L}_\textup{img}(\mathbf{I}_i, \mathbf{c}_{1,\dots,N}) = -\log \frac{\exp\left(\left\langle\Enc_{\loc}(\textbf{c}_i), \Enc_{\cont}(\textbf{I}_i)\right\rangle / \tau\right)}{\sum_{j=1}^{N} \exp\left(\left\langle\Enc_{\loc}(\textbf{c}_j), \Enc_{\cont}(\textbf{I}_{i})\right\rangle / \tau \right)}
%\end{align}
over a batch $(\mathbf{c}_i,\mathbf{I}_i)_{i=1}^N$ of $N$ coordinate-image tuples. The normalized dot-product is denoted by $\langle \cdot, \cdot \rangle$, and $\tau$ is a temperature hyperparameter.
This objective optimizes the weights of the location encoder $\Enc_{\loc}$ and image encoder $\Enc_{\cont}$ simultaneously to embed the feature vectors of the corresponding location $\Enc_{\loc}(\mathbf{c}_i) \in \mathbb{R}^d$ and image $\Enc_{\cont}(\mathbf{I}_i) \in \mathbb{R}^d$ nearby in a common $d$-dimensional feature space.

\subsection{Encoder Architectures}

Geographic \textbf{location encoders} $\Enc_{\loc}$ typically take the form
%\begin{equation}
$
    \Enc_{\loc}=\textup{NN(PE}(\mathbf{c}_{i})),
    $ \citep{Mai2023}
%\end{equation}
where $\textup{PE}(\mathbf{c}_{i})$ is a nonparametric functional positional encoding and $\textup{NN}(\cdot)$ is a small neural network. The positional encodings usually have a scale hyperparameter that controls spatial smoothness of the encoding. The neural network weights encode an implicit neural representation of a signal at a specific coordinate \citep{cole2023spatial}.
In this work, we train $\text{Siren}(\text{SH}(\mathbf{c}_{i}))$ location encoders proposed by \citet{russwurmSH2023}, which use spherical harmonics basis (SH) functions as positional encoders and are particularly well-suited for coordinates on spherical surfaces. They are combined with sinusoidal representation networks (Siren) \citep{Sitzmann2020} that are broadly used for implicit neural representations. The spatial smoothness of the representation is controlled by the number of Legendre polynomials $L$. This effectively defines the resolution of the location encoding and its capacity to learn small and large-scale geospatial patterns, with larger $L$ corresponding to finer spatial resolution.

As \textbf{image encoder}, we need a vision model that is expressive enough to learn visual patterns from satellite images. In this work, we use ResNet18, ResNet50, \citep{He2016} and ViT16 \citep{dosovitskiy2020image} vision encoders pretrained with momentum-contrast (MoCo) \citep{he2020momentum} on Sentinel-2 satellite imagery by \citet{Wang2022b}. To account for the size discrepancy between the large image models and relatively smaller location encoders (for example, the image encoder of a SatCLIP ViT16 has $\sim 22$ million parameters, while the location encoder has $\sim 1$ million parameters) during training, we freeze the vision encoder except for the last linear projection layer. 

\subsection{SatCLIP Implementation Details}
\label{sec:implementationdetails}

We pretrain \textbf{SatCLIP} using the S2-100K  dataset, which we assemble for this purpose (described in the following \cref{sec:pretraining}). We use $90\%$ of the data points, selected uniformly at random, for pretraining and reserve the remaining $10\%$ as a validation set to monitor overfitting. During pretraining, we found that batch sizes of $8k$ help the model to learn more fine-grained representations, while too large batch sizes can prevent learning, as was also recently observed on CLIP models by \citet{Zhai2023}. We train models for $500$ epochs on an A100 GPU. More pretraining details can be found in \cref{sec:SatCLIP_details}.
%\subsection{Downstream Use of SatCLIP Embeddings}
%\label{sec:fine-tuning}

\section{Experimental Setup}
\label{sec:datasets}
In our experiments, we focus on three research questions. From a performance perspective, we ask \emph{how generalizable are SatCLIP embeddings} from Sentinel-2 data, both \emph{across a diverse range of geospatial modeling tasks} (\textit{RQ 1}) and \emph{across unseen geographic areas} (\textit{RQ 2}), compared to existing pretrained location encoders and location-only prediction? We design experiments to test the performance of SatCLIP embeddings for downstream tasks, both for spatial interpolation, and for geographic domain generalization, in which the training and test sets are separated geographically. Geographic generalization is an important aspect of performance, as distributional changes across geographic areas are a common challenge in environmental problems like species distribution modeling \citep{Beery2022}, land cover classification \citep{russwurm2020meta}, crop type mapping \citep{kondmann2021denethor}, and yield prediction \citep{Tseng2022}. While location-only prediction is generally unsuitable for geographic generalization tasks, the implicit neural representation of environmental factors within the SatCLIP location encoder may provide generalizable information supporting prediction in areas with no or few labeled data points available.

After benchmarking performance, we turn to analyses that help us to further understand why SatCLIP works and what limitations it might exhibit. Ablation studies systematically vary the image encoder architecture and the scale parameters of the location encoder to understand the joint effect of each design choice. Following the intuition of \cref{fig:satclip_intuition_figure} and complementing these quantitative experiments, we ask: \emph{Do SatCLIP embeddings capture ground conditions and incorporate similarities over space?} (\textit{RQ 3}). Qualitative experiments shed light on what spatial relationships the pretrained SatCLIP embeddings harvest from multi-spectral satellite imagery. Before presenting results in \cref{sec4}, we detail our experimental design, including the pretaining dataset (\cref{sec:pretraining}), downstream tasks (\cref{sec:downstreamdata}), and comparison models (\cref{sec:comparisonmodels}).

\subsection{Pretraining Dataset: 100k uniformly sampled Sentinel-2 images}
\label{sec:pretraining}

\begin{figure}[t]
 \centering
   \includegraphics[width=\textwidth]{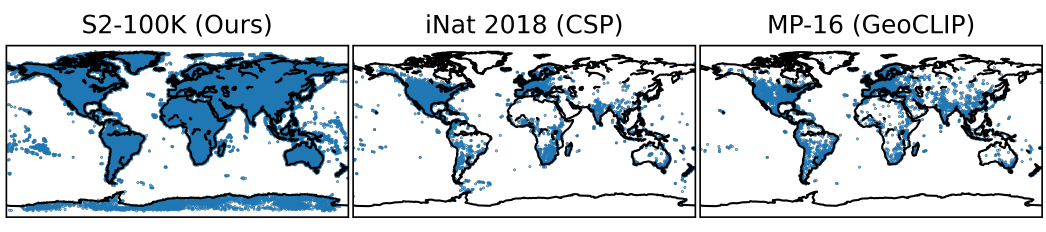}
   \caption{\textbf{Spatial distribution of the S2-100K dataset} used for training SatCLIP compared with iNaturalist 2018 \citep{Horn2018} and MP-16 \citep{Larson2017}, which are used to pretrain CSP and GeoCLIP models. iNaturalist and MP-16 heavily overrepresent North America and Europe.
   }
   \label{fig:dist}
 \end{figure}

\noindent
To construct our pretraining dataset, \textbf{S2-100K}, we sample $100,000$ tiles of $256 \times 256$ pixel, multi-spectral ($12$-channel) Sentinel-2 satellite imagery and their associated centroid locations. We design the S2-100K dataset with the goals of \textit{multi-task applicability} and \textit{geographic generalization performance} in mind. Our dataset (1) represents general location features by using multi-spectral satellite imagery (as illustrated in \cref{fig:satclip_intuition_figure}) \emph{and} (2) is nearly uniformly distributed across global land mass (\cref{fig:dist}, left). More details on the S2-100K dataset and sampling procedure are provided in \Cref{sec:s2-100k}.
In contrast, the pretraining datasets used in comparison methods (\cref{sec:comparisonmodels} and \cref{sec:competing_location_encoders}) often significantly underrepresent certain -- especially non-Western -- geographic areas, as they were not specifically designed to provide general-purpose embeddings. \cref{fig:dist} illustrates the spatial coverage of S2-100K compared to the highly clustered distributions of iNaturalist, which is used as a pretraining dataset for CSP \citep{Mai2023} and MediaEval 2016, which is used as a pretraining dataset for GeoCLIP \citep{Cepeda2023}. Similar biases are exhibited by the Yahoo-Flickr Creative Commons 100 Million (YFCC100M)  dataset \citep{Thomee2016} of image-tag-location triplets (used in GPS2Vec \citep{Yin2019}), and the Functional Map of the World (FMoW) \citep{Christie2018} dataset, which samples satellite imagery mostly near human-built infrastructure.% (also used in CSP). 

\subsection{Downstream Tasks}
\label{sec:downstreamdata}

\begin{table*}[b]
\centering\small
\caption{\textbf{Key characteristics of SatCLIP vs.\ the GPS2Vec, CSP and GeoCLIP location encoders and the MOSAIKS feature extractor.}}
\label{tab:overview}
\resizebox{\textwidth}{!}{ 
\begin{tabular}{lcc}
    \toprule
       & \textbf{Contextual data}            & \textbf{Location encoder}   \\
    \midrule
        \textbf{SatCLIP} (ours)                                & {S2-100K}                        & {Spherical harmonics \& Siren}    \\
    {\textbf{GPS2Vec} \citep{Yin2019}} 
                        & YFCC images and semantic tags                   & {Two-level soft encoding}           \\
    {\textbf{CSP} \citep{Mai2023}}                                & {FMoW / iNaturalist}               & {Sinusoidal transform \& FcNet}         \\
     {\textbf{GeoCLIP} \citep{Cepeda2023}}                             & {MediaEval 2016}                 & {Random Fourier Features \& MLP}               \\        
    {\textbf{MOSAIKS} \citep{Rolf2021}}& {Planet Basemaps 2019 Q3}                              & {N/A (direct feature extractor)}          \\
    \bottomrule
\end{tabular}
}
\end{table*}

To test the general applicability of SatCLIP location embeddings, we run experiments on a wide range of geospatial predictive modeling tasks. In all datasets, the inputs are raw latitude/longitude coordinates, which we transform into location embeddings. The nine downstream datasets used in this work span socioeconomic and environmental applications. We predict variables including \textbf{Air Temperature} \citep{Hooker2018} and \textbf{Elevation} \citep{Rolf2021} from coordinates as environmental regression objectives. To capture socioeconomic factors, we regress \textbf{Median Income} \citep{Jia2020}, \textbf{California Housing} prices \citep{KelleyPace2003}, and logged \textbf{Population Density} \citep{Rolf2021}. We additionally classify \textbf{Biomes}, \textbf{Ecoregions} \citep{dinerstein2017ecoregion}, and compile a new country code classification task \textbf{Countries}. Lastly, we classify \textbf{iNaturalist} species \citep{Horn2018}. Here, we have additional image features extracted from an InceptionV3 model released by \citep{Aodha2019}, which we concatenate with location embeddings during downstream training. Further details on downstream experiments can be found in \cref{sec:details_exp}.

\subsection{Comparison Methods}
\label{sec:comparisonmodels}

We compare trained SatCLIP location embeddings to GPS2Vec \citep{Yin2019}, CSP \citep{Mai2023} and GeoCLIP \citep{Cepeda2023} pretrained location embeddings. We refer to each comparison model by first stating the pretraining algorithm and then the pretraining dataset. For instance, CSP-FMoW represents CSP pretraining on FMoW dataset. Unless stated otherwise, we show results from SatCLIP models using a ViT-16 vision encoder.
%In our experiments, we compare the performance of our pretrained SatCLIP embeddings with other pretrained location encoders (CSP, GPS2Vec, GeoCLIP), unsupervised image features (MOSAIKS) and direct location-only prediction (Identity) at predicting the nine selected variables. More details on all downstream tasks can be found in \cref{sec:downstream}.
%
A summary of each comparison method is given in \cref{tab:overview}, with details given in \cref{sec:competing_location_encoders}. Like SatCLIP, GeoCLIP and CSP use the CLIP loss, with CSP adding loss terms for negative location sampling and SimCSE. GPS2Vec uses a KL divergence loss on the context (image and semantic tags) and location data.
% CSP \citep{Mai2023} use a sine-cosine-based positional encoding (\emph{grid} \citep{Mai2020a}) together with a 4-layer multi-layer perception (MLP) with skip connections termed \emph{FcNet}, proposed initially by \citep{Aodha2019} as location encoder, and InceptionV3 and ResNet50 models for iNat 2018 and FMoW data, respectively. 
% %
% GPS2Vec \citep{Yin2019} encodes locations by combining an exponential functional transform of each coordinate and its respective UTM-zone centroid with a simple three-layer ReLu network, and extracts text and image context features via vocabulary-based and CNN approaches.
% %
% GeoCLIP \citep{Cepeda2023} encodes coordinates using Random Fourier Features (RFF), combined with an MLP. Their image encoder is a pretrained vision transformer.
%
To compare to an image-only embedding, we use globally precomputed MOSAIKS \citep{Rolf2021} features, accessed via \url{siml.berkeley.edu} \citep{MOSAIKS_API}. 
%
%For a given latitude/longitude pair, MOSAIKS returns random convolutional features extracted from the nearest gridded satellite image. We access pre-computed MOSAIKS features from \url{siml.berkeley.edu} \citep{MOSAIKS_API}, which provides image features derived from Planet Basemaps from 2019, Quarter 3, at a gridded resolution of $0.01^\circ$.
%
To assess the performance improvement from the integration of contextual information over location-only prediction, we also compare to downstream learners trained on raw latitude/longitude coordinates $g(\mathbf{c})$. We refer to this approach as ``Identity'' throughout our experiments.
%A comparison between the key characteristics of SatCLIP and all comparison methods can be found in \cref{tab:overview}. Further details on these methods and their pretraining datasets are given in \cref{sec:competing_location_encoders}.

\subsection{Downstream Model Training}
\label{sec:downstreamtraining}

For all downstream tasks, we train multi-layer perceptron (MLP) models $g$ with location embeddings and raw latitude/longitude coordinates as input to predict a (continuous or discrete) outcome variable $y$. Regression models use a mean squared error (MSE) loss, and classification models use cross-entropy loss. Hyperparameters like learning rate, number of layers, or hidden dimensions are tuned using a random search on an independent validation set. All results are reported for an unseen test set. More details on the downstream task setups can be found in \cref{sec:details_exp}.

\section{Results}
\label{sec4}
\newcommand{\sidenote}[1]{{\hfill {\normalsize (\textit{#1})}}}

\subsection{Downstream Task Performance \sidenote{RQ 1}}
\label{sec:res:rq1}

\textbf{Quantitative comparison across downstream tasks.}
%We use the SatCLIP embeddings as input to a downstream model, which we train for each downstream task separately, as detailed in \cref{sec:implementationdetails}. 
\cref{tab:res_condensed} shows performance across different downstream tasks and methods. % (detailed in \cref{sec:comparisonmodels})
SatCLIP embeddings (in either $L=10$ or $L=40$ configuration) achieve the best prediction scores by a large margin on seven of the nine tasks. The exceptions are the Cali. Housing dataset, which is limited to California, and the Median Income dataset, which only contains data from the continental United States (US). For both tasks, the GeoCLIP model trained on US-centric MP-16 data performs equal to or better than SatCLIP. We also observe that SatCLIP embeddings with higher spatial resolution ($L=40$) perform better than coarse-grained ($L=10$) embeddings at these regionally constrained tasks.

% Generally, our evaluation setup including multiple, diverse tasks is out-of-scope for both CSP \citep{Mai2020a} and GPS2Vec \citep{Yin2019}, which are trained for in-domain deployment. Only GeoCLIP \citep{Cepeda2023} embeddings provide some predictive value. Of the existing methods, only MOSAIKS \cite{Rolf2021} investigate the generalization of their method across different tasks.  

\begin{table*}[t]
\centering
\caption{\textbf{Downstream task performance using SatCLIP (with ResNet50) vs.\ comparison location embeddings.} We report average test set $R^2$ and accuracy $\pm 1$ standard deviation across $10$ independently initialized  training runs.}
\label{tab:res_condensed}
\small
\resizebox{\textwidth}{!}{
\begin{tabular}{lrrrrrrrrr}
\toprule
 & \textbf{SatCLIP$_{L=10}$} & \textbf{SatCLIP$_{L=40}$} & \textbf{CSP} & \textbf{GPS2Vec} &  \textbf{MOSAIKS} & \textbf{GeoCLIP} & \textbf{Identity} \\
\textbf{Task} $\downarrow$ \textbf{Data} $\rightarrow$ \hfill  &  (S2-100K) &  (S2-100K) & (iNat) & (tag) & (Planet) & (MP-16) & ($y \sim g(\mathbf{c})$) \\
\midrule
\textbf{Regression}       &  $R^2$ $\uparrow$ &            &                                                          &            & & & \\
\cmidrule(lr){1-2}
Air temperature & $\mathbf{0.90 \pm 0.13}$ & $\mathbf{0.91 \pm 0.01}$  & $-0.56 \pm 0.59$ & $0.22 \pm 0.00$ & $-0.52 \pm 2.00$ & $-3.11 \pm 5.24$ & $0.82 \pm 0.16$ \\
Median income & $0.42 \pm 0.01$ & $\mathbf{0.47 \pm 0.12}$ &  $-0.01 \pm 0.02$ & $0.21 \pm 0.00$ & $0.02 \pm 0.05$ & $\mathbf{0.50 \pm 0.01}$ & $-0.84 \pm 0.94$ \\
Cali. housing & $0.35 \pm 0.04$ & $0.57 \pm 0.02$  & $-0.00 \pm 0.00$ & $0.71 \pm 0.03$ &  $0.24 \pm 0.02$ & $\mathbf{0.75 \pm 0.01}$ & $0.05 \pm 0.02$ \\
Elevation & $0.83 \pm 0.01$ & $\mathbf{0.88 \pm 0.00}$ &  $0.11 \pm 0.05$ & $0.10 \pm 0.00$  & $0.21 \pm 0.01$ & $0.83 \pm 0.00$ & $0.25 \pm 0.08$ \\
Population & $0.79 \pm 0.00$ & $\mathbf{0.82 \pm 0.00}$  & $0.36 \pm 0.11$ & $0.25 \pm 0.00$ & $0.46 \pm 0.02$ & $0.79 \pm 0.00$ & $0.46 \pm 0.03$ \\
\midrule
\textbf{Classification} & \% Accuracy $\uparrow$ & & & & & & \\
\cmidrule(lr){1-2}
Countries & $94.28 \pm 0.18$ & $\mathbf{96.00 \pm 0.14}$ & $82.11 \pm 1.72$ & $70.35 \pm 0.06$  & $76.16 \pm 0.50$ & $90.72 \pm 0.44$ & $82.94 \pm 2.23$ \\
iNaturalist & $\mathbf{65.69 \pm 0.18}$ & $\mathbf{66.22 \pm 0.40}$ & $60.47 \pm 0.56$ & $58.78 \pm 0.48$ & $56.73 \pm 0.8$ & $62.01 \pm 0.59$ & $60.83 \pm 0.53$ \\
Biome & $92.23 \pm 0.26$ & $\mathbf{94.41 \pm 0.14}$ &  $73.18 \pm 5.58$ & $69.69 \pm 0.06$  & $79.61 \pm 0.42$ & $89.57 \pm 0.45$ & $83.55 \pm 2.43$ \\
Ecoregions & $89.32 \pm 0.31$ & $\mathbf{91.67 \pm 0.15}$ &  $78.43 \pm 1.71$ & $68.46 \pm 0.06$ & $70.48 \pm 0.21$ & $84.65 \pm 0.32$ & $77.07 \pm 2.54$ \\

\bottomrule
\end{tabular}}
\end{table*}

\begin{figure}[t]
    \centering
    \includegraphics[width=1\columnwidth]{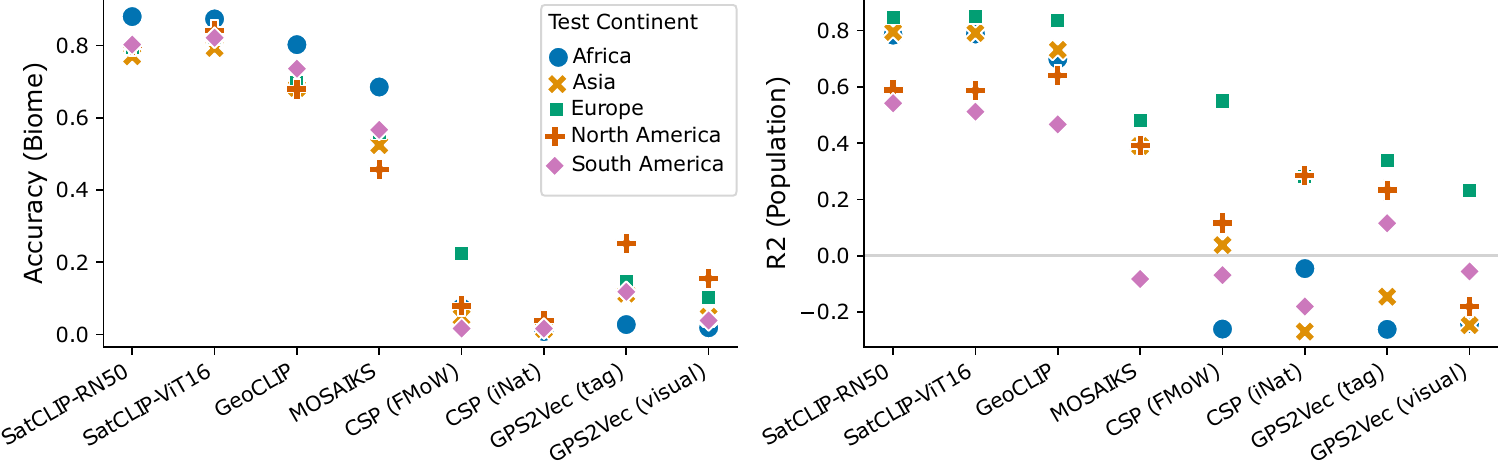}
    \caption{\textbf{Performance metrics aggregated by continent} highlight how location embeddings perform in different geographic areas for population density estimation and biome classification for five continents. $L=40$ SatCLIP models are shown.}
    \label{fig:continent_performance}
\end{figure}

\textbf{Comparison across continents.}
\cref{fig:continent_performance} shows the performance of SatCLIP and comparison methods evaluated separately by continent for the tasks of biome classification and population density estimation. 
SatCLIP performs well on all continents for both tasks.
% In terms of biome classification accuracy, SatCLIP performs well on all continents. 
Prior location encoders (CSP and GPS2Vec) trained on spatially biased training data tend to perform better in Europe and North America than in the underrepresented continents of Africa, Asia, and South America. 
%For Population estimation SatCLIP (both with ViT and ResNet vision encoders) perform well. 
GeoCLIP is closest in performance to SatCLIP for both tasks, performing similarly well to SatCLIP on the population density task, but worse across continents on the Biome classification task.

\subsection{Zero/Few-Shot Geographic Adaptation  \sidenote{RQ 2}}
\label{sec:res:rq2}

\cref{tab:res_adaptation_condensed} shows how our embeddings perform at geographic generalization. For these experiments, we deploy a spatial train/test split strategy: We hold out entire continents, either Africa or Asia, as test sets and use the remaining data for model training and validation. Since Countries and Ecoregions are often unique to a continent, a zero-shot adaptation approach without any training points on the respective test continents is not possible. Here, we provide a small portion ($1\%$, uniformly sampled) of test continent points in the training set to create a de-facto few-shot geographic adaptation setting. For the remaining tasks, we test zero-shot adaptation by not providing any training points from the held-out test continent. In iNaturalist 2018, for example, this means that the model will not be able to recognize any species that are endemic to the test continent (i.e., that do not live in any other continent) but only those known from training continents.

\begin{table*}[t]%[ht]
\centering
\caption{\textbf{Geographic adaptation capabilities of SatCLIP (with ViT16 vision encoder) vs.\ baseline location embeddings to new geographic areas with no ($*$) or very few ($\dagger$) samples  from the held-out test continent.}  We report average test set $R^2$ and accuracy in \% $\pm 1$ standard deviation across $10$ independently initialized fine-tuning runs.}
\label{tab:res_adaptation_condensed}
\footnotesize
\resizebox{\textwidth}{!}{
\begin{tabular}{lrrrrrrrrrr}
\toprule
 & \textbf{SatCLIP$_{L=10}$} & \textbf{SatCLIP$_{L=40}$} &  \textbf{CSP} & \textbf{GPS2Vec}  & \textbf{MOSAIKS} & \textbf{GeoCLIP} & \textbf{Identity} \\
\textbf{Test Continent} &  (S2-100K) &  (S2-100K) & (iNat) & (tag)  & (Planet) & (MP-16) & ($y \sim g(\mathbf{c})$) \\
\midrule
\textbf{Asia}     &  &  &            &            &               &                                           & & \\
\midrule
Air Temp.$^*$ R$^2$ $\mathbf{\uparrow}$ & $\mathbf{0.75 \pm 0.05}$ & $0.63 \pm 0.04$ &  $-0.50 \pm 1.32$ & $-3.95 \pm 4.89$ &  $-2.13 \pm 3.50$ & $\mathbf{0.77 \pm 0.28}$ & $0.20 \pm 1.64$ \\
Elevation$^*$  & $\mathbf{0.46 \pm 0.09}$ & $\mathbf{0.48 \pm 0.07}$ &  $-0.26 \pm 0.03$ & $-0.29 \pm 0.01$  & $-0.07 \pm 0.06$ & $\mathbf{0.50 \pm 0.03}$ & $-0.16 \pm 0.06$ \\
Pop. Density$^*$ & $\mathbf{0.42 \pm 0.08}$ & $\mathbf{0.45 \pm 0.04}$ &  $-1.02 \pm 0.32$ & $-0.37 \pm 0.04$  & $0.05 \pm 0.12$ & $\mathbf{0.38 \pm 0.04}$ & $0.03 \pm 0.07$ \\
\cmidrule(lr){1-1}
Countries$^\dagger$ \% Acc. $\uparrow$  & $\mathbf{36.90 \pm 4.32}$ & $19.17 \pm 2.82$ &   $1.28 \pm 0.01$ &  $1.12 \pm 0.00$ &  $1.56 \pm 0.47$ & $23.12 \pm 2.50$ & $1.24 \pm 0.12$ \\
iNaturalist$^*$  & $19.60 \pm 0.78$ & $\mathbf{20.91 \pm 0.77}$ & $\mathbf{21.49 \pm 0.85}$ & $17.52 \pm 0.38$ & $16.14 \pm 0.42$ & $\mathbf{20.94 \pm 0.38}$ & $\mathbf{21.08 \pm 0.69}$ \\
Biome$^*$      & $25.89 \pm 2.79$ & $16.44 \pm 1.21$  &  $3.00 \pm 2.60$ &  $1.76 \pm 0.04$ & $\mathbf{37.81 \pm 4.47}$ & $\mathbf{31.67 \pm 1.91}$ & $6.24 \pm 2.71$ \\
Ecoregions$^\dagger$  & $\mathbf{21.02 \pm 1.09}$ & $10.86 \pm 1.19$ &  $1.41 \pm 0.14$ &  $1.49 \pm 0.03$ &    $1.36 \pm 0.10$ & $6.65 \pm 1.03$ & $1.52 \pm 0.47$ \\
\midrule
\textbf{Africa}     &  &  &            &            &               &                                           & & \\
\midrule
Air Temp.$^*$ R$^2$ $\uparrow$ & $-4.71 \pm 2.29$ & $\mathbf{-1.48 \pm 0.70}$ &  $-2.67 \pm 5.80$ & $-7.91 \pm 0.04$  & $-17.43 \pm 18.37$ & $-9.91 \pm 28.82$ & $-27.36 \pm 39.46$ \\
Elevation$^*$ & $-1.80 \pm 1.74$ & $\mathbf{-0.21 \pm 0.09}$ & $-1.20 \pm 0.55$ & $\mathbf{-0.13 \pm 0.06}$  &  $-0.79 \pm 0.43$ & $-0.34 \pm 0.10$ & $-2.43 \pm 2.67$ \\
Pop. Density$^*$ & $0.17 \pm 0.12$ & $0.18 \pm 0.09$  & $-0.31 \pm 0.16$ & $-0.34 \pm 0.02$  &  $0.15 \pm 0.05$ & $\mathbf{0.32 \pm 0.03}$ & $-0.50 \pm 0.34$ \\
\cmidrule(lr){1-1}
Countries$^\dagger$ \% Acc. $\uparrow$ & $\mathbf{30.65 \pm 4.23}$ & $10.22 \pm 1.62$ &  $0.45 \pm 0.04$ & $0.47 \pm 0.01$ & $0.48 \pm 0.00$ & $10.32 \pm 2.75$ & $2.74 \pm 2.52$ \\
iNaturalist$^*$ & $\mathbf{9.53 \pm 0.57}$ & $6.23 \pm 0.47$ &  $8.65 \pm 0.52$ & $7.47 \pm 0.53$ & $5.18  \pm 0.38$ & $7.69 \pm 0.30$ & $\mathbf{9.96 \pm 0.33}$ \\
Biome$^*$  & $35.72 \pm 5.48$ & $12.34 \pm 1.75$ & $1.09 \pm 0.48$ & $1.29 \pm 0.04$ &  $\mathbf{49.86 \pm 1.57}$ & $28.28 \pm 3.06$ & $1.46 \pm 0.67$ \\
Ecoregions$^\dagger$ & $\mathbf{32.03 \pm 1.19}$ & $12.91 \pm 1.63$ & $0.94 \pm 0.04$ & $0.88 \pm 0.01$  & $0.92  \pm 0.12$ & $12.41 \pm 2.20$ & $7.72 \pm 3.93$ \\
\midrule
\# of wins & 8 & 5  & 1 & 0 & 2 & 4 & 2 \\
\bottomrule
\end{tabular}}
\end{table*}

\begin{figure}[ht!]
    \centering
    \includegraphics[width=1\textwidth]{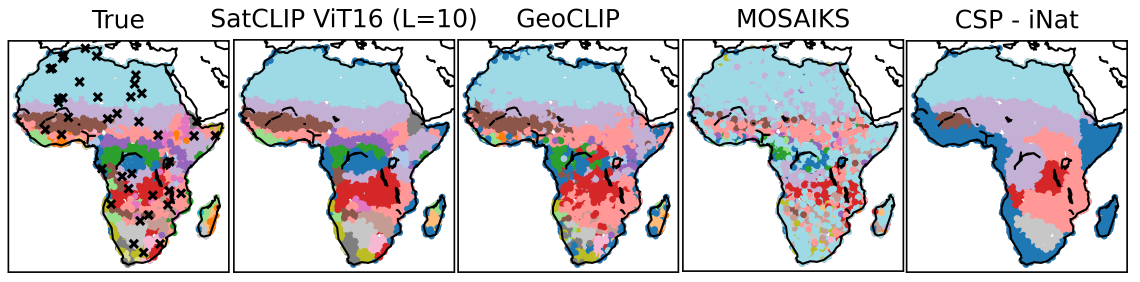}
    \caption{\textbf{Geographic adaptation: predictions of Ecoregions for Africa}. SatCLIP with $L=10$ maps Ecoregions in Africa closest to the ground truth, followed by GeoCLIP. MOSAIKS provides predictions that are too fine-grained, and CSP-iNat is too coarse. ``X'' marked locations in the ``True'' panel show the sparse training locations in Africa, which are on average 480km apart from their nearest neighbor.}
    \label{fig:ecoregions_africa}
\end{figure}

%Models trained with SatCLIP embeddings can adapt to new geographic areas with zero (tasks marked by $^*$) or very few (tasks marked by $^\dagger$) samples. 
SatCLIP models are often (but not always) better than the comparison approaches across both held-out continents. $L=10$ SatCLIP outperforms the higher-resolution $L=40$ SatCLIP model (8 vs. 5 task wins). Here, smoother representations with the lower resolution $L=10$ allow information from spatially far points to take effect. GeoCLIP is the closest to SatCLIP, with 4 wins. While 3 wins are shared with SatCLIP, it is uniquely best in population estimation in Africa. Overall, SatCLIP embeddings perform systematically better in the few-shot geographic adaptation setting for Ecoregions and Countries tasks. Several existing location encoders even perform significantly worse than directly encoding latitude and longitude for out of sample prediction (``Identity'' in \cref{tab:res_adaptation_condensed}). \cref{fig:ecoregions_africa} visualizes the predictions and failure modes from different methods on Ecoregion prediction in Africa. 
%\textcolor{red}{[todo: fill in this paragraph depending on what we end up showing with fig. 5 -- would be nice to show the distribution of fine-tuning data.]}
%
% 
% This indicates that SatCLIP is able to utilize spatial variation in Sentinel-2 imagery into the direct location encoder, in a way that improves upon simple image-agnostic representations.
%Overall, these results indicate that Sentinel-2 data contains features that can, to some degree, generalize across the planet and across application domains--and, that this information can be encoded directly into a context-pretrained location encoder.

\subsection{Analysis of Location Embeddings \sidenote{RQ 3}}
\label{sec:res:rq3}
\begin{figure}[t]%[htb]
    \begin{subfigure}{1\linewidth}
    \centering
    \includegraphics[width=0.95\linewidth]{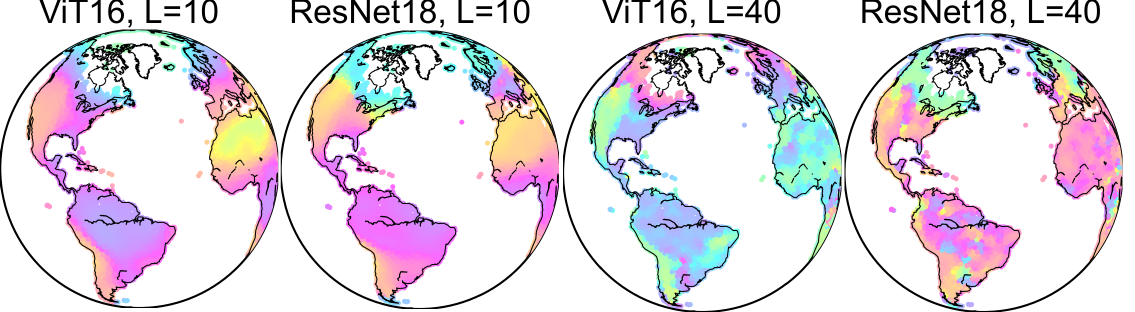}
    \caption{\textbf{Latent space visualization}: The first $3$ principal components from the $256$-dimensional SatCLIP embedding as RGB for four models with different $L$ values and vision encoders. PCAs are calculated individually for each globe, and colors across globes are incomparable. 
    %Across vision encoders, the same patterns are learned, such as different PCAs for the Amazon rainforest and Andes mountains or the US East and West Coast. By design, location encoders L=10 produce smoother representations than L=40.
    }
    \label{fig:res:pca}
    \end{subfigure}
    \begin{subfigure}{\linewidth}

        \centering
        \includegraphics[width=1\textwidth]{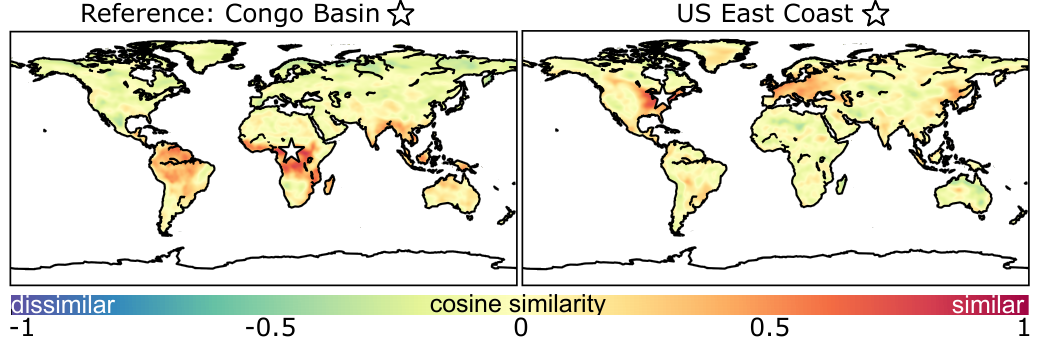}
        %\input{figures/embeddings_legend.tikz}
        %\caption{\textbf{Analysis of environmental factors learned} by the trained SatCLIP location encoder by a T-SNE \citep{van2008visualizing} dimensionality reduction at globally distributed points colored by biome type. Overall, biomes are clearly separated in this T-SNE embedding space. Note that some points of the northern $\triangle$ and southern $\triangledown$ hemispheres are mapped nearby, which indicates that this clustering is due to the trained SatCLIP weights and not the pure coordinates, as we investigate further in the appendix \cref{fig:appendix:embeddings}.}
        \caption{\textbf{Location similarities learned by SatCLIP} give insights into areas with similar visual features. Highlighted are similarities between location embeddings for a Congo Basin location (left) and a US East Coast location (right)--marked by a star--and the rest of the world. SatCLIP associates the Congo Basin with other Equatorial locations, e.g, in the Amazon, and associates the US East Coast with other densely populated areas, e.g., in Central Europe.}
        \label{fig:locationsim}
    \end{subfigure}
    \caption{\textbf{Analysis of SatCLIP location embeddings} through (a) low dimensional PCA projection as RGB visualizations and (b) cosine similarities of different locations of the full $256$-dimensional embedding space.}
\end{figure}

We now investigate qualitatively to what degree the SatCLIP embeddings have learned an implicit representation of different ground conditions in the location encoder weights. We first visualize a low-dimensional projection of the latent representations learned by our location encoders. \cref{fig:res:pca} shows an RGB representation of the first three principal components of SatCLIP embeddings at locations around the planet. The figure highlights how embeddings learned by SatCLIP provide fine-grained representations of different locations (expressed by different colors), capturing global patterns like climate zones. The figure also highlights the different spatial smoothness of $L=10$ and $L=40$ SatCLIP embeddings and changes between models using different vision encoders.

Next, we examine similarities between embeddings of different locations measured in the cosine distance between the embedding of a location $\Enc_{\loc}(\mathbf{c})$ with respect to a reference location $\Enc_{\loc}(\mathbf{c}_\star)$. \cref{fig:locationsim} shows the similarity of a grid of locations, i.e., the map with respect to reference locations in the Congo Basin and on the east coast of North America on the right panel. The reference locations are marked by a star $\star$ on the map. SatCLIP location embeddings show high similarity between the Congo Basin location and other areas close to the Equator, particularly the Amazon and Indonesia (red areas on the left panel). In comparison, embeddings of the North American location are similar to areas in Europe or northern China that are similarly population-dense and industrialized.

\subsection{Effect of Location and Image Encoder Design on Performance}

Lastly, we report what effect the vision encoder type and spatial resolution of the location encoder ($L$ hyperparameter) have on the downstream accuracy. The choice of image encoder for pretraining (ViT-16, ResNet-18, ResNet-50) appears to only marginally affect results (performance differences of $<1\%$). In contrast, different location encoder resolutions with scale parameters $L=10$ and $L=40$ have a greater effect: the $L=40$ location encoders tend to be better for interpolation tasks, as shown in  \cref{tab:modelcomparison}. The $L=10$ location encoders are better for the zero- and few-shot adaptation experiments (not shown in the main paper, but in \cref{appendix:full_res}). %, as outlined in the full tables in appendix \cref{appendix:full_res}, augmenting \cref{tab:res,tab:res_adaptation}.
% These figures are meant to provide an intuition for what SatCLIP embeddings represent, but it is important to acknowledge that the full $256$-dimensional SatCLIP embeddings capture many more subtleties than can be visualized here, as we discuss in \Cref{sec:appendix:embedding}.

\begin{table}[t]
\caption{Comparisons of different vision encoders (\% accuracy) in spatial interpolation classification tasks (comp. \cref{tab:res_condensed}). Full results on all tasks can be found in \cref{appendix:full_res}.
%Within each smoothness level L=10 or L=40, SatCLIP embeddings with all vision models achieve very similar accuracies on downstream tasks within 1-$\sigma$. Hence, all vision models are able to extract location-specific patterns.
}

    \label{tab:modelcomparison}
    \centering
\footnotesize
\resizebox{\textwidth}{!}{
\begin{tabular}{lcccccc}%rrrrrr
\toprule
                   & ViT16,L=10        & ResNet18,L=10      & ResNet50,L=10     & ViT16,L=40         & ResNet18,L=40      & ResNet50,L=40      \\
                    \cmidrule(lr){2-2}  \cmidrule(lr){3-3}  \cmidrule(lr){4-4}   \cmidrule(lr){5-5}   \cmidrule(lr){6-6}  \cmidrule(lr){7-7}    
Countries          & $93.97 \pm 0.30$  & $93.92 \pm 0.30$   & $94.28 \pm 0.18$  & $ 95.77 \pm 0.14$  & $ 95.92 \pm 0.10$  & ${96.00 \pm 0.14}$ \\
iNaturalist        & $65.69 \pm 0.50$  & $65.56 \pm 0.29$   & $65.50 \pm 0.43$  & ${65.98 \pm 0.61}$ & ${66.40 \pm 0.49}$ & ${66.03 \pm 0.54}$ \\
Biomes             & $92.07 \pm 0.22$  & $92.10 \pm 0.23$   & $92.23 \pm 0.26$  & ${94.27 \pm 0.15}$ & ${94.33 \pm 0.10}$ & ${94.41 \pm 0.14}$ \\
Ecoregions         & $89.53 \pm 0.28$  & $89.57 \pm 0.23$   & $89.32 \pm 0.31$  & ${91.61 \pm 0.22}$ & ${91.53 \pm 0.15}$ & ${91.67 \pm 0.15}$ \\
\bottomrule
\end{tabular}}

\end{table}

\section{Discussion}
\label{sec5}
%\textcolor{red}{[ER note: We can either write a discussion section here that summarizes and discusses the key results, or we can put these sentences below back in the results sections. I prefer to state results in results, then discuss results in discussion, but I can see a case for either way. I don't think either is more costly in terms of space, so we should think of flow.]}
% Discussion section
In reference to {RQ1}, the results presented in \cref{sec:res:rq1} show that \emph{SatCLIP models can provide useful information for a wide range of downstream tasks}, with the concrete benefit varying across tasks. SatCLIP outperformed other approaches on globally distributed downstream tasks, while for regional datasets such as Cali. Housing, other methods like GeoCLIP were competitive. SatCLIP embeddings perform well across continents and are less prone to geographic bias in comparison to other methods like GPS2Vec or CSP, where performance degrades outside of Europe or North America.

Regarding \emph{RQ2}, the results in \cref{sec:res:rq2} indicate that the \emph{transfer of spatial patterns in Sentinel-2 imagery into the SatCLIP location encoder enables generalization across geographic areas}. SatCLIP is the best of all models tested under conditions of geographic domain generalization and generally improves upon our satellite image-only (``MOSAIKS'') and location-only (``Identity'') baseline methods. T
%his indicates that during pre-training, spatial variation in imagery is translated to the location encoder. 
%At the same time, the remaining spatial smoothness in the location embeddings facilitates interpolation between pre-training locations during test time.  
At test time, SatCLIP location encoders can be applied directly to any point on the globe, without needing to download additional imagery.
Additional experiments showed that the downstream performance of our current SatCLIP is more affected by changes to the location encoder scale factor than the exact vision architecture used for the image encoder. This could relate to the relative model sizes--the vision architectures we use here are much larger than the location encoder architectures. It also suggests that when designing future iterations of SatCLIP and similar models, improvements in vision and location encoder design as well as their balancing during training should be explored.
%
% most global in-sample tasks, having a higher-resolution (less smooth) embedding surface (L=40 in our location encoder) was generally better on average on our downstream tasks, though still a smoother (L=10) representation was performing best on some tasks, i.e., iNaturalist. 
% %
% For out-of-domain generalization, smoother representations were generally preferred.
%
Finally, \emph{qualitative analyses support these interpretations of our results}.
Our similarity analysis of location embeddings in \cref{sec:res:rq3} showed that spatially far locations like in South America (Amazon rainforest) and Africa (Congo Basin) are embedded nearby in the SatCLIP embedding space due to the visual similarity of satellite images from these locations. This can explain why SatCLIP models generalize to unseen geographic areas with no or little training data, as demonstrated in our geographic domain generalization experiments. %, which is a key challenge in geospatial machine learning.

\section{Conclusion}
We presented a method to learn an implicit neural representation of visual patterns on the globe by matching satellite images and their respective coordinates using a contrastive location-image pretraining objective: SatCLIP. 
Experiments show that SatCLIP is effective for global prediction tasks spanning social and environmental domains, for both interpolation and out-of-sample geographic prediction, and compared to existing location encoders, image-only and location-only prediction.
The effectiveness of SatCLIP is complemented by its relative simplicity in implementation (a single contrastive loss on $100,000$ openly available satellite images). 
Two key factors contribute to the performance observed in \cref{sec:res:rq1,sec:res:rq2}. First, while our S2-100K pretraining dataset is smaller than the pretraining datasets used for other geographic location encoders, our samples are \emph{uniformly distributed across the globe} in a way that supports globally distributed downstream tasks. 
Second, we use the recently proposed Siren(SH) location encoder, which has proven well-suited for the global-scale representation of data on the spherical Earth.
% This simple but effective SatCLIP framework lays the foundation for a new class of models: \emph{global-coverage, general-purpose location encoders} that coalesce large, geo-tagged data into a succinct representation of any location.

Our findings motivate several avenues for potential future work. 
First, a key limitation of the current implementation of SatCLIP is that it uses Sentinel-2 satellite imagery as the sole source of contextual grounding for our location encoder. In fact, the image encoder we utilize in SatCLIP can be seen as a special case of a more general \textsl{context} encoder that may integrate other location-specific data modalities like audio from acoustic sensors or text from geolocated social media posts for multi-source geospatial learning. 
Second, the current SatCLIP pre-trained weights have limited spatial scales, dictated by the $L$ parameter of the location encoder.
For extremely high resolution or local phenomena, it will be important to further study the effect of different choices of the location encoder and its use for downstream learning.
Third, there is the potential to expand the SatCLIP training framework to encode locations in both time and space. Here, we effectively marginalized over all time points in the S2-100k dataset (which is sampled over two years and thus includes seasonal differences in images) but did not directly embed time in a space-time encoder of e.g., the form $f(lat,lon,time)$.
Addressing each of these three limitations would require separate innovations (on e.g.model architecture), and would need a new set pretraining data and/or downstream tasks for evaluation. Thus, we see these extension as deserving of their own dedicated study in future work. 
Code for SatCLIP pretraining and downstream experiments as well as the S2-100K dataset is available at \url{github.com/microsoft/satclip}.

{
    \small
    \bibliographystyle{ieeenat_fullname}
    \bibliography{konstantin}
}

\clearpage
\setcounter{page}{1}
%\maketitlesupplementary
\appendix  

\section{S2-100K Dataset Overview} 
\label{sec:s2-100k}

We sample the S2-100k dataset from all Sentinel 2 Level-2A scenes that meet the following criteria:
\begin{enumerate}
    \item Are captured between January 1st, 2021 and May 17th, 2023
    \item Are estimated to have less than 20\% cloud cover (as reported by the European Space Agency preprocessing)
    \item Are at least partially over a land mass (measured with country level boundaries from \url{gadm.org})
\end{enumerate}

Using the Microsoft Planetary Computer, we find $2,359,972$ S2-L2A scenes that meet these criteria. We sample a random scene, sample a random $256 \times 256$ pixel patch from that scene, then keep the patch if less than 10\% of the pixels include nodata values, else we reject the patch. We repeat this process until we have $100,000$ patches. Each patch contains all available bands -- B01, B02, B03, B04, B06, B06, B07, B08, B08A, B09, B11, and B12 -- resampled to a 10m/px spatial resolution in the native UTM coordinate system and saved as a single Cloud Optimized GeoTIFF. Finally, we record the latitude and longitude (using the EPSG:4326 coordinate system) of the center point of the patch. 
The dataset can be downloaded at \url{github.com/microsoft/satclip}. 
The final dataset is $83.79$ Gigabytes on disk.

\section{Further Details on Other Geographic Location Encoders \& Their Pretraining Datasets}
\label{sec:competing_location_encoders}

\subsection{(Pre-)training Data}

Here, we describe the datasets used for pretraining in previous geographic location encoders, CSP and GPS2Vec, in detail:

\textbf{iNat 2018} \citep{Horn2018}: This dataset contains crowdsourced, natural images of over $8,000$ plant and animal species from around the globe. The iNat 2018 training set contains over $400,000$ image-location pairs. While species distributions can be indicative of e.g., climate zones \citep{Aodha2019}, they are less descriptive of socio-economic processes, as we show in our experiments. Furthermore, most iNat image locations lie in North America and Europe, with only few observations in other continents, as we show in \cref{fig:dist}. We obtain location encoders pretrained on iNat from CSP \citep{Mai2023}.

\textbf{FMoW} \citep{Christie2018}: This dataset contains satellite images annotated with bounding boxes representing $63$ categories describing functional or land-use characteristics (e.g., ``airport'' or ``gas station''). The dataset contains over $363,000$ image-location pairs. FMoW is aimed to support detection of functional objects from satellite imagery and, as such, is skewed towards human-built infrastructure and includes fewer natural scenes. It is also again heavily favoring Western countries in its geographic distribution. We obtain location encoders pretrained on FMoW from CSP \cite{Mai2023}.

\textbf{YFCC100M} \citep{Thomee2016}: This dataset contains natural images and associated semantic tags collected from the social media platform Flickr. YFCC100M includes approximately $48$ million image-tag-location triplets. Images (and tags) represent diverse scenes, from street views to birthday parties. While they are more indicative of social dimensions and physical infrastructure, YFCC100M imagery is less representative of natural processes. Just like iNat and FMoW, the dataset again mostly contains locations in Western countries. We obtain models separately trained on images and semantic tags from GPS2Vec \cite{Yin2019}.

\textbf{MP-16} \citep{Larson2017}: The MediaEval Placing Tasks 2016 (MP-16) dataset also consists of images from Flickr. Overall, it includes $4.27$ million images and their corresponding geo-tag. Like the YFCC100M dataset used for training GPS2Vec models, MP-16 is geographically clustered in areas with higher Flickr activity and overrepresents Western countries.

\subsection{CSP}

\textbf{Location and image encoders}: The authors use a \textbf{location encoder} combining sinusoidal transforms introduced by \citep{Mai2020a} with a fully-connected neural network. The \textbf{image encoder} is dataset-dependent. For the iNat dataset, the authors use a pretrained IncpetionV3 network. For the FMoW dataset, they use a pretrained ResNet50 network. In both cases, neural network weights in all layers except the last linear projection layer are frozen. 

\textbf{Learning objective}: Here, the authors combine (1) the standard CLIP loss, which leverages in-batch negative sampling and which is also used by us, with two other contrastive objectives: (2) Random negative location sampling (i.e. predicting the real location from a set of randomly generated locations) and (3) SimCSE sampling (i.e. matching location embeddings obtained using different dropout masks). The two extra objectives help the model to balance the location and context encoders. We find that the two additional objectives are not needed for training SatCLIP.

\subsection{GPS2Vec}

\textbf{Location and context encoders}: The authors train \textbf{location encoders} specific to the UTM zones containing training data. They first encode coordinates as an exponential function of the Euclidean distance between a latitude/longitude coordinate and its respective UTM zone centroid coordinate. This is followed by a simple ReLu network with three hidden layers. GPS2Vec does not learn a \textbf{context encoder} but directly extracts text features using a vocabulary-based approach and image features using convolutional neural networks (CNN). 

\textbf{Learning objective}: The authors design an objective which aims to use contextual features (images and semantic tags) as labels to train their location encoder by estimating the normalized frequency of features in the vicinity of a given location. Practically, this is achieved by minimizing the KL-divergence of context and location embedding distributions. 

\subsection{GeoCLIP}

\textbf{Location and image encoders}: The \textbf{location encoder} in GeoCLIP first transforms raw latitude/longitude coordinates into equal earth projection (EEP), then extracts features using Random Fourier Features (RFF). On top of this positional encoding, the authors deploy separate MLPs for each RFF which are aggregated into the final location embedding. For their \textbf{image encoder}, the authors use a CLIP pretrained ViT16 models. Just like for SatCLIP, the image encoder is frozen except the last two linear projections layers. 

\textbf{Learning objective}: GeoCLIP models are trained using the CLIP objective, outlined in \cref{eq:cliploss}. The only difference to SatCLIP training is that the authors add additional location negatives (randomly sampled) to each batch. 

\section{SatCLIP (Pre)-training} 
\label{sec:SatCLIP_details}

\subsection{Training Details}
\label{sec:SatCLIP_details:training}

\noindent \textbf{Batch size:}  After experimenting with different batch sizes, we opt for models trained at batch sizes of $8k$. While traditional CLIP image-text pretraining behaves optimally at a batch size of $32k$ \citep{Zhai2023}, we observe most effective training at $8k$ and $16k$. 

\noindent \textbf{Image encoder:} We train SatCLIP models with ViT16, ResNet18 and ResNet50 image encoders, all pretrained on Sentinel-2 imagery and published by \cite{Wang2022a}. We keep the image encoders frozen during training, and only train a last projection layer that maps the image embeddings into the desired output space. We find this to be ideal for training at a size of $256$--this is equivalent to the embedding size used by CSP \cite{Mai2023}. 

\noindent \textbf{Location encoder:} Our location encoder follows an approach recently proposed by \cite{russwurmSH2023}. It first puts raw latitude/longitude coordinates through a functional transform based on orthogonal spherical harmonics. The number of spherical harmonics used corresponds to the resolution at which the model is trained. In practice this is controlled by a parameter defining the number of Legendre polynomials $L$ used for spherical harmonics computation. We train a lower-resolution ($L=10$) and a higher-resolution ($L=40$) version of SatCLIP. The functional transform is followed by a sinusoidal representation network (Siren) consisting of two hidden layers and $512$ hidden dimensions. These hyperparameters are obtained after rigurously testing different settings.

\noindent \textbf{Augmentations:} We deploy several data augmentations within our training procedure. Image augmentations include random crops, random horizontal flipping, random vertical flipping and Gaussian blurs. Point coordinates are augmented using a coordinate jitter which randomly shifts image coordinates by up to about 1km.

\noindent \textbf{Optimization:} All SatCLIP models are trained with the Adam optimizer, a learning rate of $0.0001$ and a relatively high weight-decay values of $0.01$ to help prevent overfitting. All final SatCLIP models are trained for $500$ epochs. On our single A100 GPU, training takes around $2$ days. Throughout training, we reserve $10\%$ of the data for validation. We monitor validation loss and select SatCLIP models according to the minimum validation loss.

\subsection{Scale Sensitivity}
\label{sec:SatCLIP_details:scale}

\begin{figure}
    \centering
    \includegraphics[width=0.5\textwidth]{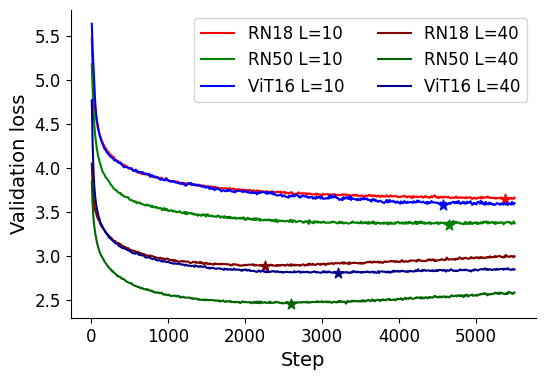}
    \caption{Validation set loss of SatCLIP models trained with different vision encoder backbones (ResNet18, ResNet50 and ViT16) with $L=10$ and $L=40$. Stars indicate the minimum validation loss, which we use for model selection.}
    \label{fig:loss}
\end{figure}

We want to briefly comment on the sensitivity of SatCLIP training to the number of legendre polynomials $L$ used in the model's spherical harmonics location encoder. This hyperparameter effectively controls the spatial resolution of the obtained embeddings. Smaller values of $L$ are computationally more efficient and ideal for representing large-scale patterns, while larger values of $L$ are better for capturing small-scale patterns. More details on this can be found in \citet{russwurmSH2023}. During SatCLIP training, we observe some interesting differences between smaller ($L=10$) and higher ($L=40$) resolution models. We find that higher-resolution models are more likely to exhibit overfitting, as \cref{fig:loss} highlights. On downstream tasks, higher resolution models perform better at smaller scale, regional tasks as e.g. the Cali. Housing dataset, as \cref{tab:appendix:satclipmodels} highlights. Lastly, $L=40$ models appear better for spatial interpolation (RQ1), while $L=10$ models seem better suited for geographic generalization (RQ2).

\section{Latent Space Exploration}
\label{sec:appendix:latent}

\subsection{SatCLIP Embeddings in Different Biomes}
\label{sec:appendix:embedding}

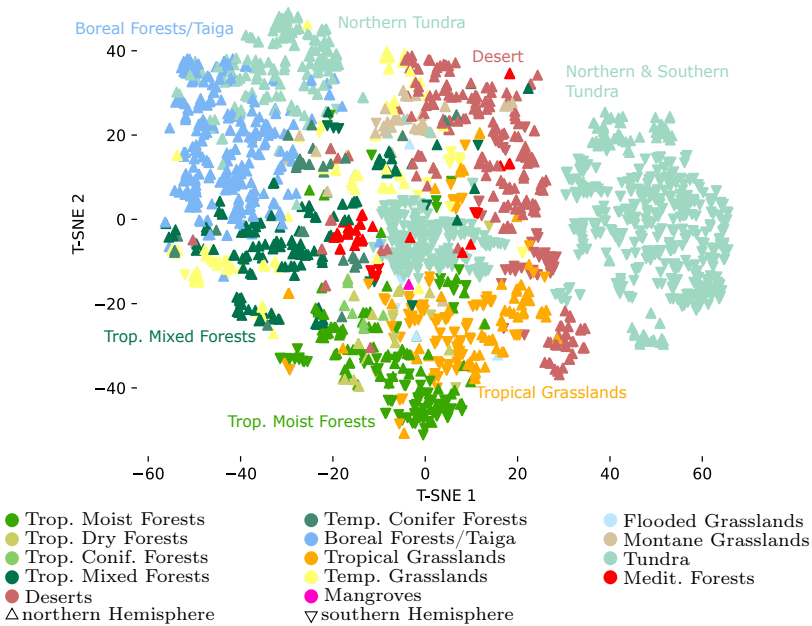
\begin{figure}[ht]%[htb]
        \centering
        \includegraphics[width=.75\linewidth]{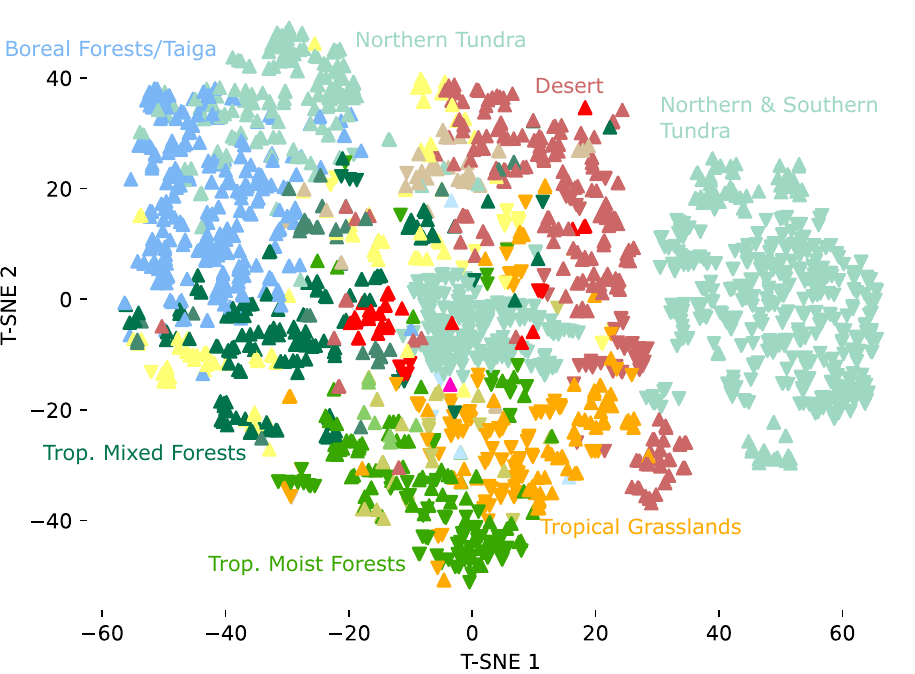}
        \begin{tikzpicture}[yscale=-0.25, 
xscale=4, 
font=\scriptsize, 
inner sep=0.2em,
triangle/.style = {fill=blue!20, regular polygon, regular polygon sides=3 },
node rotated/.style = {rotate=180},
border rotated/.style = {shape border rotate=180}
]

\node[circle, fill=TropicalSubtropicalMoistBroadleafForests, label={right:Trop. Moist Forests}] at (0,0){};
\node[circle, fill=TropicalSubtropicalDryBroadleafForests, label={right:Trop. Dry Forests}] at (0,1){};
\node[circle, fill=TropicalSubtropicalConiferousForests, label={right:Trop. Conif. Forests}] at (0,2){};
\node[circle, fill=TemperateBroadleafMixedForests, label={right:Trop. Mixed Forests}] at (0,3){};
\node[circle, fill=TemperateConiferForests, label={right:Temp. Conifer Forests}] at (1,0){};
\node[circle, fill=BorealForestsTaiga, label={right:Boreal Forests/Taiga}] at (1,1){};
\node[circle, fill=TropicalSubtropicalGrasslandsSavannasShrublands, label={right:Tropical Grasslands}] at (1,2){};
\node[circle, fill=TemperateGrasslandsSavannasShrublands, label={right:Temp. Grasslands}] at (1,3){};
\node[circle, fill=FloodedGrasslandsSavannas, label={right:Flooded Grasslands}] at (2,0){};
\node[circle, fill=MontaneGrasslandsShrublands, label={right:Montane Grasslands}] at (2,1){};
\node[circle, fill=Tundra, label={right:Tundra}] at (2,2){};
\node[circle, fill=MediterraneanForestsWoodlandsScrub, label={right:Medit. Forests}] at (2,3){};
\node[circle, fill=DesertsXericShrublands, label={right:Deserts}] at (0,4){};
\node[circle, fill=Mangroves, label={right:Mangroves}] at (1,4){};
{};
\node[triangle, draw=black, fill=none, label={right:northern Hemisphere\phantom{p}}, inner sep=0.1em](a) at (0,5){};
\node[triangle, border rotated, draw=black, fill=none, label={right:southern Hemisphere}, inner sep=0.1em]at (1,5){};
\end{tikzpicture}
        \caption{\textbf{Analysis of environmental factors learned} by the trained SatCLIP location encoder by a T-SNE \citep{van2008visualizing} dimensionality reduction at globally distributed points colored by biome type. Overall, biomes are clearly separated in this T-SNE embedding space. Note that some points of the northern $\triangle$ and southern $\triangledown$ hemispheres are mapped nearby, which indicates that this clustering is due to the trained SatCLIP weights and not the pure coordinates, as we investigate further in the appendix \cref{fig:appendix:embeddings}.}
        \label{fig:tsne:groundconditions}
\end{figure}

\begin{figure*}
    \begin{subfigure}{.32\textwidth}
        \includegraphics[width=\textwidth]{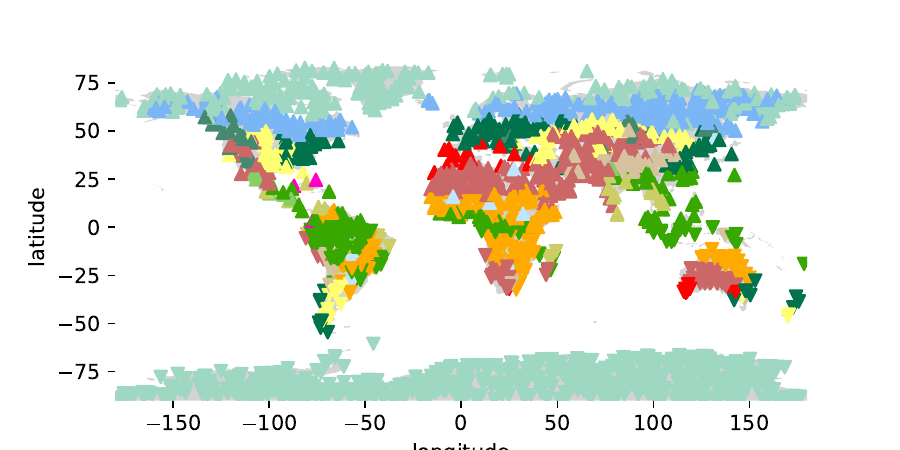}
        \caption{Map (lon/lat) representation of all biomes}
        \label{fig:appendix:embeddings:full:latlon}
    \end{subfigure}
    \begin{subfigure}{.32\textwidth}
        \includegraphics[width=\textwidth]{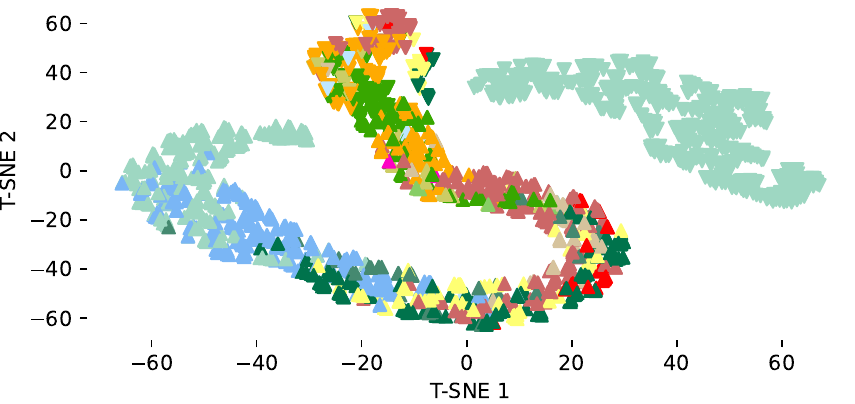}
        \caption{Spherical harm. (SH) embedding (no weights).}
        \label{fig:appendix:embeddings:full:pe}
    \end{subfigure}
    \begin{subfigure}{.32\textwidth}
        \includegraphics[width=\textwidth]{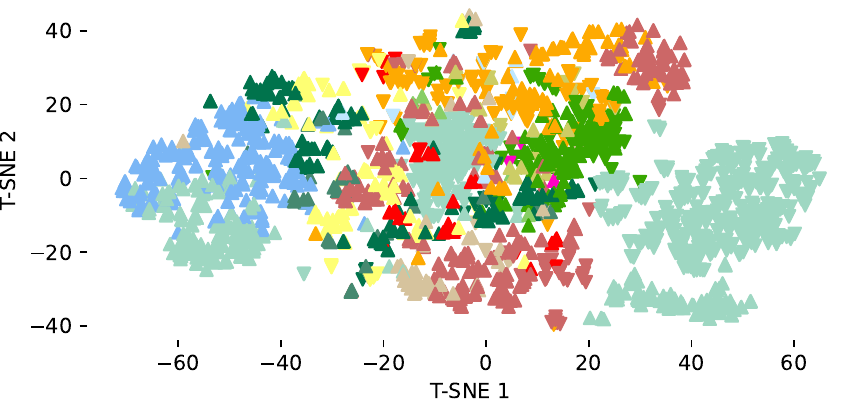}
        \caption{SatCLIP (SH \& trained SirenNet)}
        \label{fig:appendix:embeddings:full:SatCLIP}
    \end{subfigure}
    
    \begin{subfigure}{.32\textwidth}
        \includegraphics[width=\textwidth]{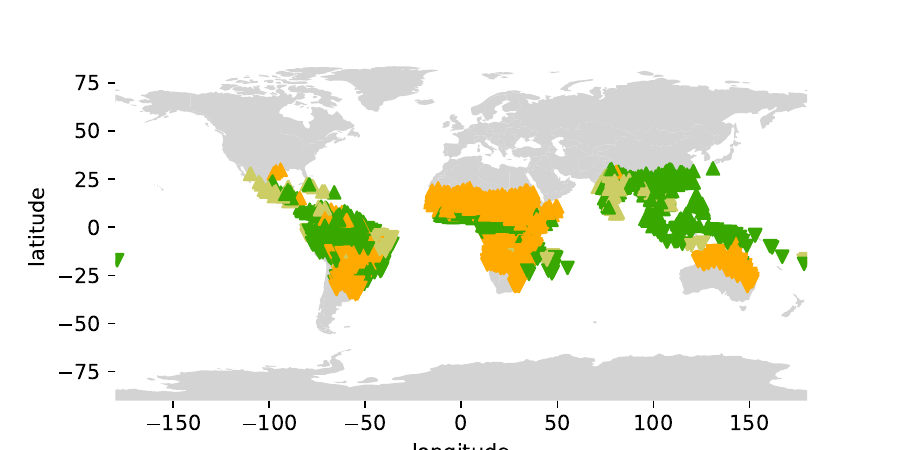}
        \caption{Difficult-to-separate tropical biomes}
        \label{fig:appendix:embeddings:selected:latlon}
    \end{subfigure}
    \begin{subfigure}{.32\textwidth}
        \includegraphics[width=\textwidth]{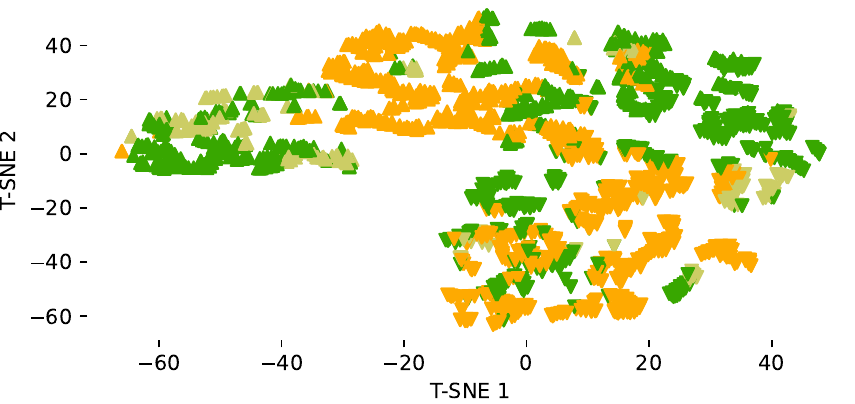}
        \caption{Spherical harm. (SH) embedding (no weights).}
        \label{fig:appendix:embeddings:selected:pe}
    \end{subfigure}
    \begin{subfigure}{.32\textwidth}
        \includegraphics[width=\textwidth]{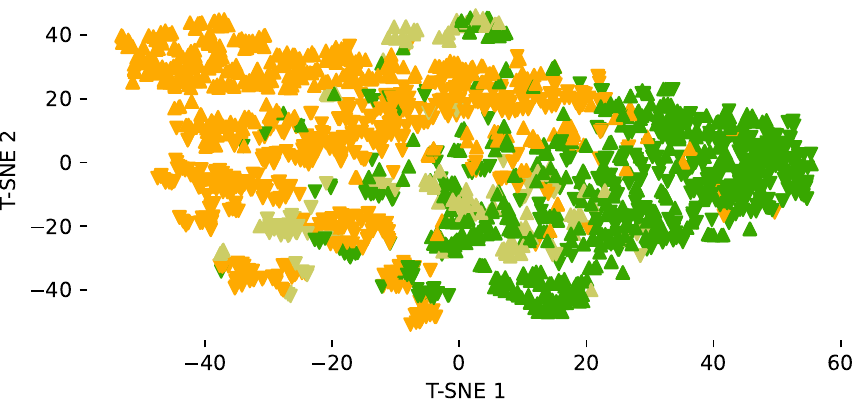}
        \caption{SatCLIP (SH \& trained SirenNet)}
        \label{fig:appendix:embeddings:selected:SatCLIP}
    \end{subfigure}

    \begin{subfigure}{0.9\textwidth}
    \centering
        \definecolor{watercolor}{HTML}{0000FF}
\definecolor{TropicalSubtropicalMoistBroadleafForests}{HTML}{38A700}
\definecolor{TropicalSubtropicalDryBroadleafForests}{HTML}{CCCD65}
\definecolor{TropicalSubtropicalConiferousForests}{HTML}{88CE66}
\definecolor{TemperateBroadleafMixedForests}{HTML}{00734C}
\definecolor{TemperateConiferForests}{HTML}{458970}
\definecolor{BorealForestsTaiga}{HTML}{7AB6F5}
\definecolor{TropicalSubtropicalGrasslandsSavannasShrublands}{HTML}{FEAA01}
\definecolor{TemperateGrasslandsSavannasShrublands}{HTML}{FEFF73}
\definecolor{FloodedGrasslandsSavannas}{HTML}{BEE7FF}
\definecolor{MontaneGrasslandsShrublands}{HTML}{D6C39D}
\definecolor{Tundra}{HTML}{9ED7C2}
\definecolor{MediterraneanForestsWoodlandsScrub}{HTML}{FE0000}
\definecolor{DesertsXericShrublands}{HTML}{CC6767}
\definecolor{Mangroves}{HTML}{FE01C4}

\begin{tikzpicture}[yscale=-0.3, 
xscale=5.5, 
font=\scriptsize,
inner sep=0.2em,
triangle/.style = {fill=blue!20, regular polygon, regular polygon sides=3 },
node rotated/.style = {rotate=180},
border rotated/.style = {shape border rotate=180}
]

% First column
\node[circle, fill=TropicalSubtropicalMoistBroadleafForests, label={right:(Sub-)Tropical Moist Broadleaf Forests}] at (0,0){};
\node[circle, fill=TemperateConiferForests, label={right:Temperate Conifer Forests}] at (0,1){};
\node[circle, fill=TropicalSubtropicalConiferousForests, label={right:(Sub-)Tropical Coniferous Forests}] at (0,2){};
\node[circle, fill=TemperateBroadleafMixedForests, label={right:Temperate Broadleaf \& Mixed Forests}] at (0,3){};
\node[circle, fill=BorealForestsTaiga, label={right:Boreal Forests/Taiga}] at (0,4){};
\node[circle, fill=FloodedGrasslandsSavannas, label={right:Flooded Grasslands \& Savannas}] at (0,5){};
\node[circle, fill=DesertsXericShrublands, label={right:Deserts \& Xeric Shrublands}] at (0,6){};
\node[circle, fill=Mangroves, label={right:Mangroves}] at (0,7){};

% Second column
\node[circle, fill=TropicalSubtropicalGrasslandsSavannasShrublands, label={right:(Sub-)Tropical Grasslands, Savannas \& Shrublands}] at (1,0){};
\node[circle, fill=TropicalSubtropicalDryBroadleafForests, label={right:(Sub-)Tropical Dry Broadleaf Forests}] at (1,1){};
\node[circle, fill=TemperateGrasslandsSavannasShrublands, label={right:Temperate Grasslands, Savannas \& Shrublands}] at (1,2){};
\node[circle, fill=MontaneGrasslandsShrublands, label={right:Montane Grasslands \& Shrublands}] at (1,3){};
\node[circle, fill=Tundra, label={right:Tundra}] at (1,4){};
\node[circle, fill=MediterraneanForestsWoodlandsScrub, label={right:Mediterranean Forests, Woodlands \& Scrub}] at (1,5){};
\node[triangle, border rotated, draw=black, fill=none, label={right:southern Hemisphere}] at (1,6){};
\node[triangle, draw=black, fill=none, label={right:northern Hemisphere\phantom{p}}]at (1,7){};

\end{tikzpicture}

    \end{subfigure}
    
    \centering
    \caption{Supplementary experiment to \cref{fig:tsne:groundconditions}: we disentangle the influence of spherical harmonic (SH) coordinate embedding without trainable weights in Figs b) \& e) with the trained SatCLIP location encoder in Figs c) \& f). Comparing the T-SNE embedding projections in these two columns reveals the additional effect of having trained SatCLIP weights int he location encoder. 
    We perform this analysis in an easier setting with all biomes (top row) and with only tropical geographically intertwined biomes (bottom row). In the top row, also the pure spherical harmonic basis functions produce a T-SNE manifold along which biomes are located b). In the more difficult bottom row, no coherent manifolds form in e) along biome types. These appear only in combination with the SatCLIP-trained weights in f)}
    \label{fig:appendix:embeddings}
\end{figure*}

This supplementary experiment disentangles the influence of raw coordinates and learned SatCLIP embeddings in the T-SNE embedding experiment of \cref{fig:tsne:groundconditions} in the main paper that indicated that the SatCLIP \emph{weights} encode environmental factors like dryness or temperature in biomes. Critically speaking, also the direct embedding of coordinates \emph{without any weights} could generate T-SNE clusters that look similar to \cref{fig:tsne:groundconditions}. In \cref{fig:appendix:embeddings}, we investigate if this is the case by showing the embeddings of coordinates with Spherical Harmonics (SH) (no trainable weights) alone (\cref{fig:appendix:embeddings:full:pe,fig:appendix:embeddings:selected:pe}) and compare it with the trained SatCLIP location encoder, which combines Spherical Harmonics basis functions with Sinusoidal Representation Networks (SirenNets) \citep{Sitzmann2020} (\cref{fig:appendix:embeddings:full:SatCLIP,fig:appendix:embeddings:selected:SatCLIP}).

In the top row \cref{fig:appendix:embeddings:full:SatCLIP,fig:appendix:embeddings:full:pe,fig:appendix:embeddings:full:latlon} we show all biomes and can see that the T-SNE points embedded in spherical harmonic basis functions of pure coordinates (without trainable weights) (\cref{fig:appendix:embeddings:full:pe}) creates a clear manifold along which the different biomes cluster. However, notice that points from the northern hemisphere ($\triangle$) are clearly separated from the southern hemisphere ($\triangledown$). Also, the arctic and antarctic tundra biomes are separated by two clusters. This is the T-SNE embedding produced from pure coordinate embeddings, that serves as the baseline to the SatCLIP embeddings that are produced by representing point coordinates as spherical harmonic basis functions \emph{and transforming these basis functions with the SatCLIP-trained SirenNet neural network}. The result is shown in \cref{fig:appendix:embeddings:full:SatCLIP}, which corresponds to \cref{fig:tsne:groundconditions} in the main paper, where the embedding clusters along dryness and temperature dimensions rather than pure geographic location. Hence, by comparing \cref{fig:appendix:embeddings:full:pe} (no weights) with \cref{fig:appendix:embeddings:full:SatCLIP}, we can see the effect of trainable SatCLIP weights on the T-SNE representation of the points.

To make sure that this analysis holds, we repeat the embedding in a more challenging setting: we show embeddings of points from only tropical biomes for (Sub-)Tropical Moist Broadleaf Forests, (Sub-)Tropical Dry Broadleaf Forests, and (Sub-)Tropical Grasslands, Savannas \& Shrublands in \cref{fig:appendix:embeddings:selected:SatCLIP,fig:appendix:embeddings:selected:pe,fig:appendix:embeddings:selected:latlon} (second row). Points from these biomes are not separable using only non-parametric embeddings (\cref{fig:appendix:embeddings:selected:pe}), since these biomes are all tropical and geographically intertwined. Here only the geographic coordinate is not sufficiently expressive to form T-SNE manifolds along biome lines. However, with the pretrained SatCLIP location encoder (\cref{fig:appendix:embeddings:selected:SatCLIP}), we can clearly identify a dry to wet trend along T-SNE dimension 1 (left to right) where points from grassland/savanna transitions first into dry broadleaf forest and then moist broadleaf forest.

In summary, these supplementary analyses reinforce the conclusions of the main paper experiment around \cref{fig:tsne:groundconditions}, which experimentally supports the concept that the SatCLIP embeddings represent an implicit neural representation of environmental and societal ground conditions that are visible in the Sentinel-2 images that the SatCLIP model is trained on.

\subsection{Latent Space Visualization at Different Scales}
\label{sec:appendix:latent_viz}

\begin{figure}[ht]
    \centering
    \includegraphics[width=1\columnwidth]{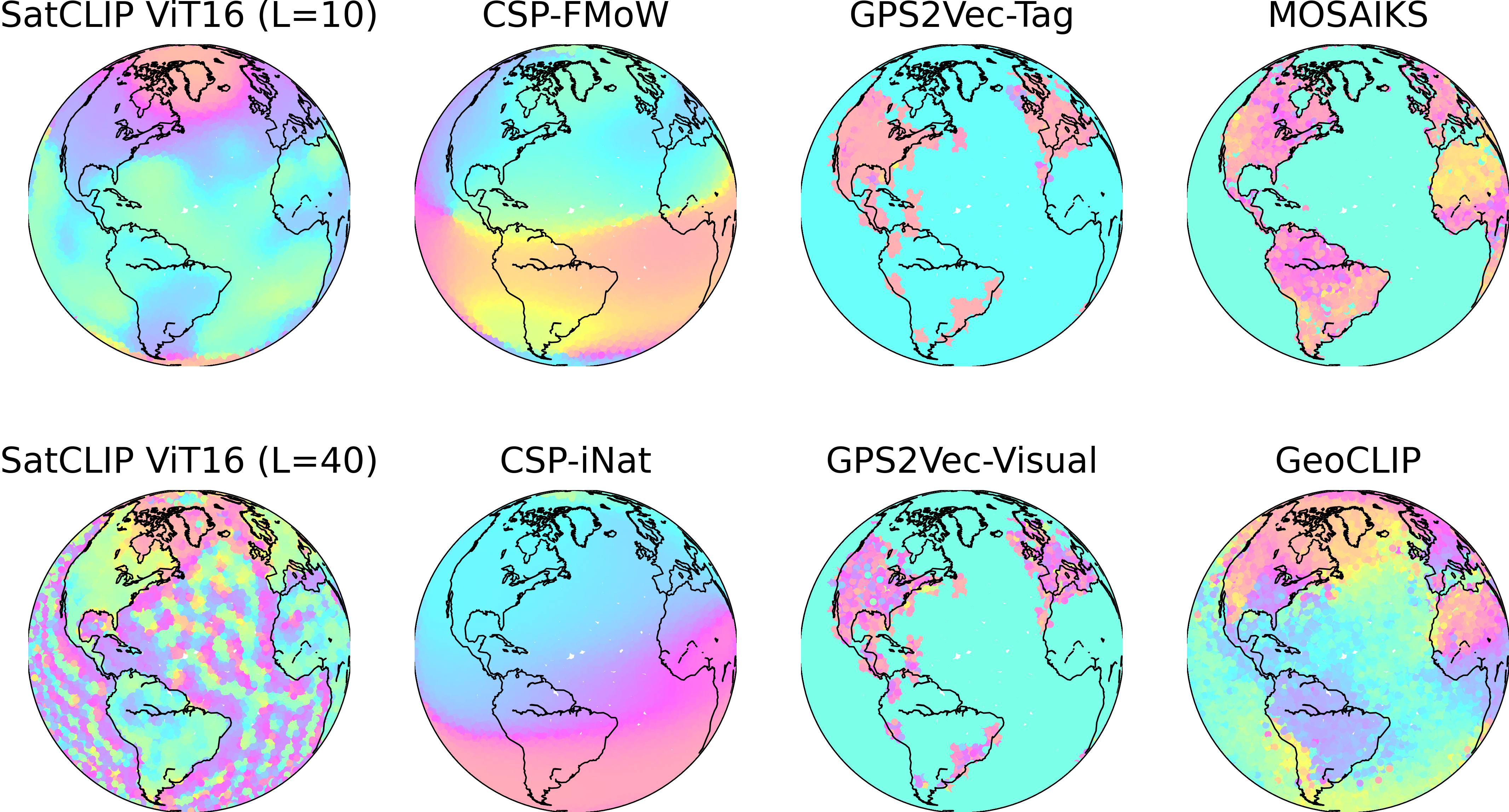}
    \caption{Visualization of the top-$3$ principal components, plotted as RGB channels, of SatCLIP models trained with different $L$ hyperparameters and comparison methods.}
    \label{fig:pca_SatCLIPs}
\end{figure}

\begin{figure*}[ht]
    \centering
    \includegraphics[scale=0.4]{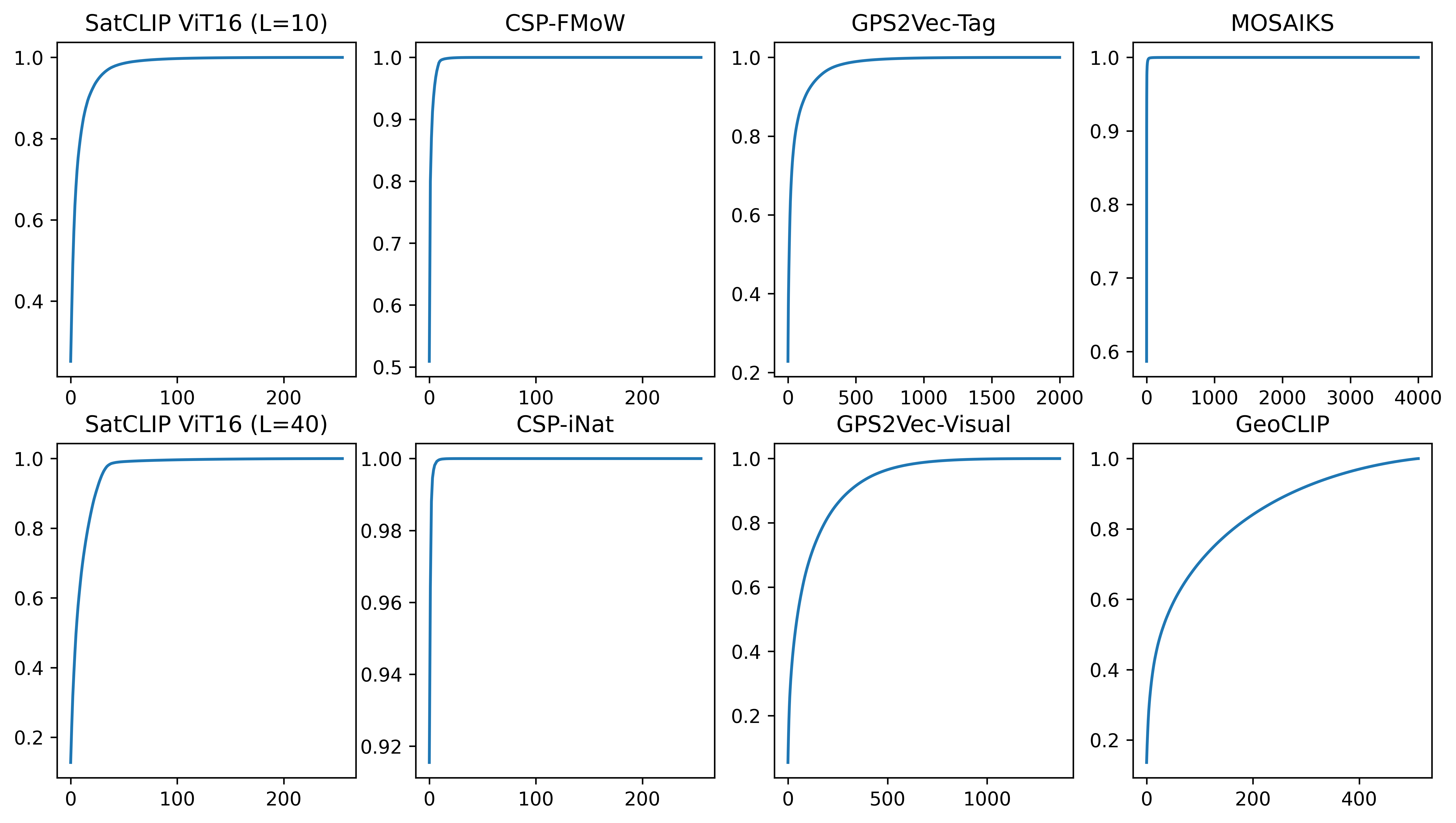}
    \caption{Curves of the explained variance ratio of a principal component analysis (PCA) conducted on the different embeddings using globally distributed locations from the Population dataset.}
    \label{fig:pca_curves}
\end{figure*}

Lastly, we provide a visualization of the embeddings learned by SatCLIP models at different scales (different $L$ values) in \cref{fig:pca_SatCLIPs}. It is clearly visible how SatCLIP $L=40$ resolves positional embeddings much more fine-grained. Lastly, in \cref{fig:pca_curves} we show the explained variance ratio by element in a respective embedding vector, extracted via PCA, including SatCLIPs with different $L$ values and comparison methods.

\begin{table*}[]
\centering
\caption{Description of the datasets used in downstream experiments.}
\label{tab:data}
\resizebox{\textwidth}{!}{%
\small
\begin{tabular}{lcccccccc}
\toprule
\textbf{Name} & \textbf{$n$} & \textbf{Spatial coverage} & \textbf{Locations} & \textbf{Outcome variable} & \textbf{Inputs} & \textbf{Task} \\ \midrule
Air Temp. \citep{Hooker2018} & $3,076$ & Global & Weather stations & Ann. mean temp. & lon, lat & Regr. \\
Med. Income \citep{Jia2020} & $3,107$ & Cont., USA & Census tract & Med. house inc. & lon, lat & Regr. \\
Cali. Housing \citep{KelleyPace2003} & $20,640$ & Calif., USA & House locations & House price & lon, lat & Regr. \\
Elev. \citep{Rolf2021} & $99,995$  & Global & Reg. sampled & Elevation & lon, lat & Regr. \\
Pop. Dens. \citep{Rolf2021} & $74,512$  & Global & Reg. sampled & Pop. dens. & lon, lat & Regr. \\
Countries (Ours) & $100,000$ & Global   & Reg. sampled & Country code & lon, lat & Class. \\
Biome  \citep{dinerstein2017ecoregion} & $100,000$ & Global      & Reg. sampled & Biome type & lon, lat & Class. \\
Ecoregions  \citep{dinerstein2017ecoregion} & $100,000$ & Global & Reg. sampled & Ecoregion & lon, lat & Class. \\
iNat 2018 \citep{Horn2018} & $460,406$  & Global & Image locations & Species classes & lon, lat, image & Class. \\ \bottomrule
\end{tabular}%
}
\end{table*}

\section{Details and Additional Results for Downstream Experiments} 
\label{sec:details_exp}

\subsection{Countries, Ecoregions, and Biome Dataset Overview}
\label{sec:countrieseco_biome}

To create the Countries, Ecoregions and Biome datasets used in Section \ref{sec4} we sample a dataset of 100,000 latitude and longitude points approximate uniformly at random, then record which country, ecoregion and biome each point falls within, or whether the point is over an ocean. We use country boundaries from the 4.1 release of ``The Database of Global Administrative Areas'' (GADM, \url{gadm.org}) for country codes. We rely on \citet{dinerstein2017ecoregion} for ecoregion and biome maps. The Countries dataset contains points over 184 countries (out of a total of 263 countries in the GADM dataset), resulting in a classification task with 185 categories (including an ``ocean'' class). The Ecoregions dataset contains 719 classes, also including an ``ocean'' class. The Biome dataset contains 15 classes and also includes an ``ocean'' class. Note that the latitude/longitude locations for all three datasets are the same.

\subsection{Downstream Task Overview}
\label{sec:downstream}

\cref{tab:data} highlights the different datasets used in downstream tasks throughout our experiments along with characteristics like their spatial coverage, outcome variables and task types. Generally, our tasks can be split into regression and (multi-)classification tasks. Our inputs are always raw longitude/latitude coordinate pairs which are processed by our pretrained location encoders to obtain location embeddings which are then used as inputs for downstream learners. The population density and elevation datasets are subsampled from the global datasets from \citet{Rolf2021}.  Only the iNat 2018 task contains additional data: image embeddings obtained via a pretrained InceptionV3 network.

\subsection{Training details}
\label{sec:downstream-details}

We tune all downstream task models before final training runs. All reported downstream experiments use simple MLP models. We tune the following hyperparameters by random search: number of hidden layers, hidden dimensions, learning rate, weight decay. The best setting is selected according to a validation loss. We then train $10$ models until convergence (as measured on the validation loss) and report mean and standard deviations of performance metrics ($R^2$, MSE or accuracy). All tuning and training is conducted on a single A100 GPU. 

For experiments reported in \cref{sec:res:rq1}, we deploy random train/val/test splits. For the Air Temp. dataset the training size is $60\%$, for the Cali. Housing and Med. Income datasets it is $50\%$ and for all other datasets it is $30\%$. The validation set size is $20\%$ for Air Temp., Cali. Housing and Med. Income and $10\%$ for all other datasets. Note that the iNat 2018 dataset comes with predefined train and test sets. Here we reserve $10\%$ of the train set for validation. For experiments reported in \cref{sec:res:rq2} train and test set are defined by the location of points within or outside of the respective test set continent. Here, we reserve $20\%$ of the training set for validation for Air Temp., Cali. Housing and Med. Income and $10\%$ for all other datasets. For Countries and Ecoregions we allocated a small portion ($1\%$) of test set points for training to allow the models to learn classes that only exist on the test set continent.

\subsection{Results of Different SatCLIP Configurations} 
\label{appendix:full_res}

We provide full experimental results for pretrained SatCLIPs (with different numbers of Legendre polynomials $L$ and different vision backbones ResNet18, ResNet50 and ViT16) in \cref{tab:appendix:satclipmodels}.

\begin{table*}[]
\caption{\textbf{SatCLIP models with different vision encoders and location encoders win varying resolutions controlled by the L parameter (maximum number of Legendre polynomial degrees).} 
We report average test set MSE and accuracy $\pm 1$ standard deviation across $10$ independently initialized MLP training runs.}\label{tab:appendix:satclipmodels}
\centering
\begin{subtable}{\textwidth}
\caption{Classification Accuracy in \% and Regression MSE scores.}
\centering
\footnotesize
\resizebox{\textwidth}{!}{
\begin{tabular}{lrrrrrr}
\toprule
Vision Encoder & ViT16 & ViT16 & ResNet18 & ResNet18 & ResNet50 & ResNet50 \\
Spherical Harm. $L$ & 10 & 40 & 10 & 40 & 10 & 40 \\
\midrule
\textbf{Regression} MSE  $\downarrow$ \\
\midrule
Air Temp.          & $\mathbf{0.20 \pm 0.31}$ & $0.25 \pm 0.02$ & $0.59 \pm 0.75$ & $0.81 \pm 0.77$ & $0.29 \pm 0.38$ & $0.27 \pm 0.03$ \\
Median Income      & $0.92 \pm 0.24$ & $\mathbf{0.67 \pm 0.01}$ & $0.87 \pm 0.22$ & $\mathbf{0.67 \pm 0.01}$ & $0.78 \pm 0.02$ & $0.71 \pm 0.16$ \\
Cali Housing       & $3.55 \pm 0.06$ & $2.62 \pm 0.28$ & $3.66 \pm 0.13$ & $\mathbf{2.46 \pm 0.08}$ & $3.68 \pm 0.23$ & $\mathbf{2.42 \pm 0.12}$ \\
Elevation          & $0.22 \pm 0.01$ & $\mathbf{0.15 \pm 0.01}$ & $0.20 \pm 0.01$ & $\mathbf{0.15 \pm 0.00}$ & $0.21 \pm 0.01$ & $\mathbf{0.15 \pm 0.00}$ \\
Population Density & $0.56 \pm 0.01$ & $0.50 \pm 0.02$ & $0.55 \pm 0.01$ & $\mathbf{0.48 \pm 0.01}$ & $0.55 \pm 0.01$ & $\mathbf{0.48 \pm 0.01}$ \\
\midrule
\textbf{Classification} \% Acc. $\uparrow$ \\
\midrule
Countries          & $93.97 \pm 0.30$ & $95.77 \pm 0.14$ & $93.92 \pm 0.30$ & $95.92 \pm 0.10$ & $94.28 \pm 0.18$ & $\mathbf{96.00 \pm 0.14}$ \\
iNaturalist        & $65.69 \pm 0.50$ & ${65.98 \pm 0.61}$ & $65.56 \pm 0.29$ & $\mathbf{66.40 \pm 0.49}$ & $65.50 \pm 0.43$ & ${66.03 \pm 0.54}$ \\
Biomes             & $92.07 \pm 0.22$ & $\mathbf{94.27 \pm 0.15}$ & $92.10 \pm 0.23$ & $\mathbf{94.33 \pm 0.10}$ & $92.23 \pm 0.26$ & $\mathbf{94.41 \pm 0.14}$ \\
Ecoregions         & $89.53 \pm 0.28$ & $\mathbf{91.61 \pm 0.22}$ & $89.57 \pm 0.23$ & $\mathbf{91.53 \pm 0.15}$ & $89.32 \pm 0.31$ & $\mathbf{91.67 \pm 0.15}$ \\
\bottomrule
\end{tabular}}
\end{subtable}
\vspace{1em}

\begin{subtable}{\textwidth}
\centering
\caption{Held-out countries regression and classification tasks.}
\resizebox{\textwidth}{!}{
\begin{tabular}{lrrrrrr}
\toprule
Vision Encoder & ViT16 & ViT16 & ResNet18 & ResNet18 & ResNet50 & ResNet50 \\
Spherical Harm. $L$ & 10 & 40 & 10 & 40 & 10 & 40 \\
\midrule
\textbf{Asia}               \\
\midrule
Air Temp.$^*$ MSE  $\downarrow$   & $\mathbf{0.85 \pm 0.17}$ & $1.26 \pm 0.15$ & $1.22 \pm 0.56$ & $1.56 \pm 0.14$ & $1.17 \pm 0.52$ & $1.50 \pm 0.10$ \\
Elevation$^*$                                     & $2.13 \pm 0.37$ & $\mathbf{2.06} \pm 0.28$ & $2.36 \pm 0.21$ & $2.58 \pm 0.26$ & $2.22 \pm 0.35$ & $3.28 \pm 0.09$ \\
Population Density$^*$                            & $2.03 \pm 0.26$ & $\mathbf{1.94 \pm 0.15}$ & $2.18 \pm 0.21$ & $\mathbf{1.92 \pm 0.23}$ & $2.06 \pm 0.21$ & $2.82 \pm 0.22$ \\
\cmidrule(lr){1-1}
Countries$^\dagger$ \% Acc. $\uparrow$            & $36.90 \pm 4.32$ & $19.17 \pm 2.82$ & $37.15 \pm 3.28$ & $13.71 \pm 3.26$ & $\mathbf{40.36 \pm 3.21}$ & $14.29 \pm 1.62$ \\
iNaturalist$^*$                                   & $19.60 \pm 0.78$ & $\mathbf{20.91 \pm 0.77}$ & $\mathbf{20.75 \pm 1.00}$ & $19.67 \pm 0.97$ & $19.75 \pm 0.49$ & $17.67 \pm 0.32$ \\
Biome$^*$                                         & $25.89 \pm 2.79$ & $16.44 \pm 1.21$ & $21.97 \pm 3.50$ & $15.02 \pm 0.97$ & $21.83 \pm 3.44$ & $\mathbf{30.26 \pm 3.00}$ \\
Ecoregions$^\dagger$                              & $\mathbf{21.02 \pm 1.09}$ & $10.86 \pm 1.19$ & $\mathbf{20.54 \pm 1.55}$ & $8.87  \pm 1.70$ & $\mathbf{20.32 \pm 1.41}$ & $8.46  \pm 0.79$ \\
\midrule
\textbf{Africa}     \\
\midrule
Air Temp.$^*$ MSE  $\downarrow$   & $4.13 \pm 1.66$ & $\mathbf{1.79 \pm 0.50}$ & $3.36 \pm 0.91$ & $17.60 \pm 25.27$ & $6.56 \pm 1.53$ & $2.17 \pm 0.33$ \\
Elevation$^*$                                     & $1.34 \pm 0.83$ & $\mathbf{0.57 \pm 0.04}$ & $1.25 \pm 0.36$ & $0.73  \pm 0.05$ & $1.41 \pm 0.51$ & $0.81  \pm 0.06$ \\
Pop. D.$^*$                                       & $\mathbf{1.97 \pm 0.29}$ & $\mathbf{1.96 \pm 0.22}$ & $2.39 \pm 0.46$ & $2.14  \pm 0.25$ & $2.63 \pm 0.50$ & $2.99  \pm 0.23$ \\
\cmidrule(lr){1-1}
Countries$^\dagger$ \% Acc. $\uparrow$            & $30.65 \pm 4.23$ & $10.22 \pm 1.62$ & $\mathbf{34.41 \pm 3.18}$ & $9.96  \pm 1.56$ & $\mathbf{35.30 \pm 2.00}$ & $8.95  \pm 1.04$ \\
iNaturalist$^*$                                   & $9.53  \pm 0.57$ & $6.23  \pm 0.47$ & $8.96  \pm 0.88$ & $6.68  \pm 0.44$ & $\mathbf{9.90  \pm 0.52}$ & $5.22  \pm 0.26$ \\
Biome$^*$                                         & $35.72 \pm 5.48$ & $12.34 \pm 1.75$ & $34.58 \pm 5.14$ & $20.60 \pm 1.40$ & $\mathbf{39.93 \pm 4.41}$ & $33.77 \pm 2.69$ \\
Ecoregions$^\dagger$                              & $\mathbf{32.03 \pm 1.19}$ & $12.91 \pm 1.63$ & $\mathbf{31.66 \pm 2.29}$ & $12.87 \pm 2.47$ & $\mathbf{32.69 \pm 1.79}$ & $13.54 \pm 2.06$ \\
\bottomrule
\multicolumn{5}{l}{\footnotesize{$^*$ Denotes zero-shot, $^\dagger$ denotes few-shot domain generalization tasks.}} 
\end{tabular}}
\end{subtable}
\end{table*}

\subsection{Additional Results: Predictive Performance}

\begin{figure*}[ht]
\centering
    \begin{subfigure}{.99\textwidth}   
            \centering
            \includegraphics[width=.99\textwidth]{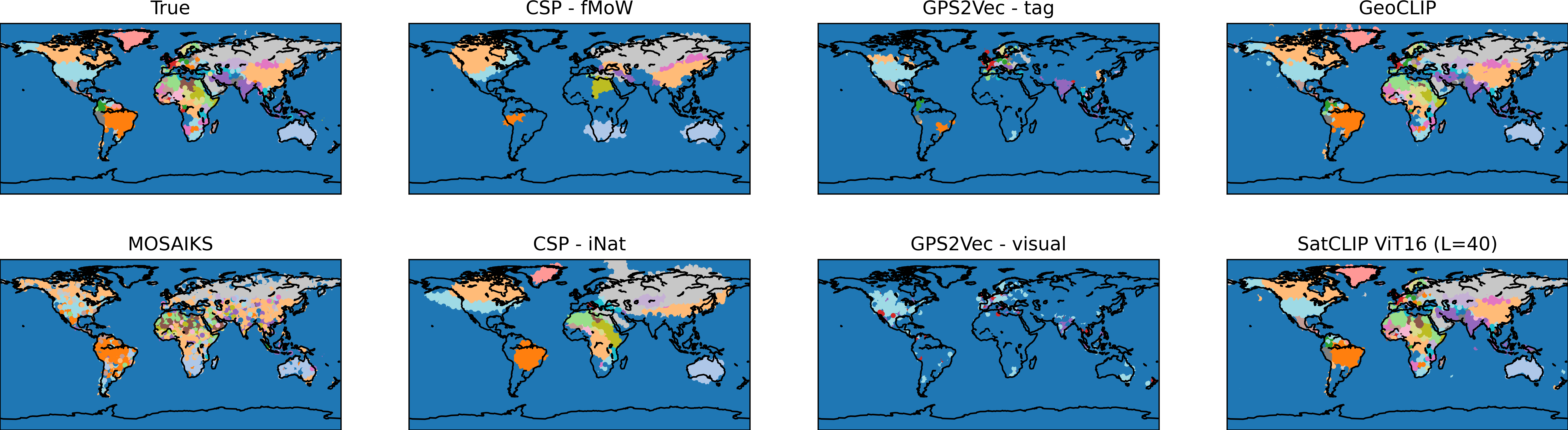}
            \caption{\textbf{Predictive performance}: Country code classification}
            \label{fig:res:countries}
    \end{subfigure}
    \hfill
    \begin{subfigure}{.99\textwidth}   
        \centering
        \includegraphics[width=.99\textwidth]{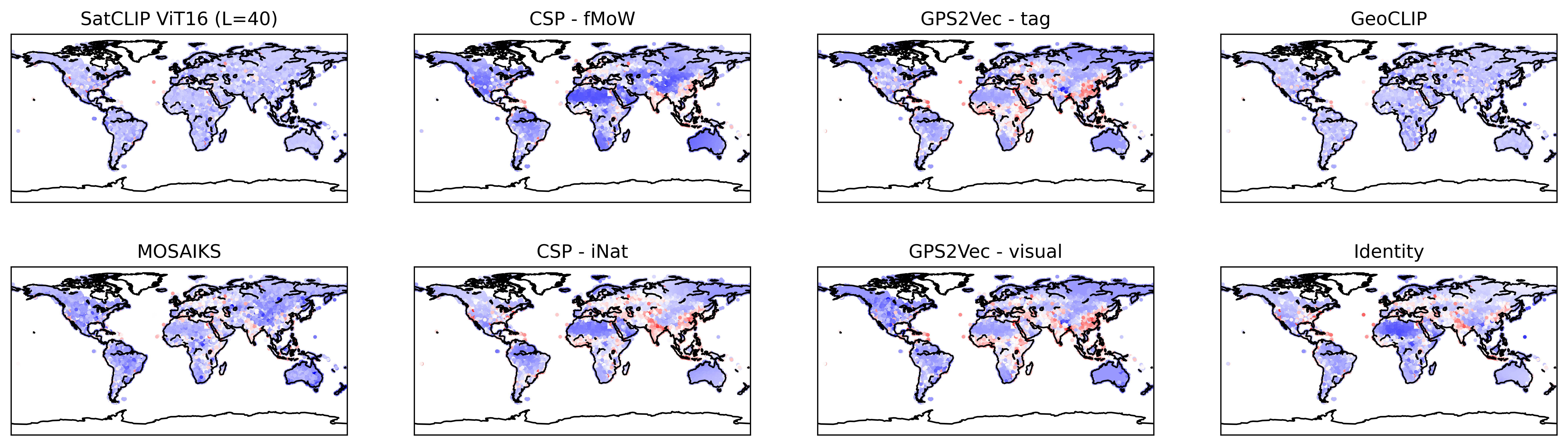}
        \caption{\textbf{Prediction errors:} Population density.} 
        \label{fig:res:population}
    \end{subfigure}
    \hfill
    \caption{ 
    \cref{fig:res:countries} shows results from the Countries dataset. Shown are predicted country codes across the planet, compared to the true country codes. \cref{fig:res:population} shows test set results from the Population dataset. Shown are model residuals (errors), red values indicate positive, blue values negative errors. Errors are standardized across figures.
    }
    \label{fig:res:full}
\vskip-0.1in
\end{figure*}

\begin{table*}[b]
\centering
\caption{\textbf{Downstream task performance using SatCLIP (with ViT16 vision encoder) vs.\ baseline location embeddings.} We report average test set $R^2$ and accuracy $\pm 1$ standard deviation across $10$ independently initialized MLP training runs.}
\label{tab:res_all}
\resizebox{\textwidth}{!}{
\begin{tabular}{lrrrrrrrrrrr}
\toprule
 & \textbf{SatCLIP$_{L=10}$} & \textbf{SatCLIP$_{L=40}$} & \textbf{CSP} & \textbf{CSP} & \textbf{GPS2Vec} & \textbf{GPS2Vec} & \textbf{MOSAIKS} & \textbf{GeoCLIP} & \textbf{Identity} \\
\textbf{Task} $\downarrow$ \textbf{Data} $\rightarrow$ \hfill  &  (S2-100K) &  (S2-100K) & (FMoW) & (iNat) & (tag) & (visual) & (Planet) & (MP-16) & ($y \sim g(\mathbf{c})$) \\
\midrule
\textbf{Regression}       &  $R^2$ $\uparrow$ &            &            &               &                               &            & & & \\
\cmidrule(lr){1-2}
Air temperature & $\mathbf{0.90 \pm 0.13}$ & $\mathbf{0.91 \pm 0.01}$ & $0.07 \pm 0.37$ & $-0.56 \pm 0.59$ & $0.22 \pm 0.00$ & $0.03 \pm 0.00$ & $-0.52 \pm 2.00$ & $-3.11 \pm 5.24$ & $0.82 \pm 0.16$ \\
Median income & $0.42 \pm 0.01$ & $\mathbf{0.47 \pm 0.12}$ & $-0.03 \pm 0.05$ & $-0.01 \pm 0.02$ & $0.21 \pm 0.00$ & $0.03 \pm 0.00$ & $0.02 \pm 0.05$ & $\mathbf{0.50 \pm 0.01}$ & $-0.84 \pm 0.94$ \\
Cali. housing & $0.35 \pm 0.04$ & $0.57 \pm 0.02$ & $0.00 \pm 0.00$ & $-0.00 \pm 0.00$ & $0.71 \pm 0.03$ & $0.61 \pm 0.03$ & $0.24 \pm 0.02$ & $\mathbf{0.75 \pm 0.01}$ & $0.05 \pm 0.02$ \\
Elevation & $0.83 \pm 0.01$ & $\mathbf{0.88 \pm 0.00}$ & $0.36 \pm 0.04$ & $0.11 \pm 0.05$ & $0.10 \pm 0.00$ & $0.06 \pm 0.00$ & $0.21 \pm 0.01$ & $0.83 \pm 0.00$ & $0.25 \pm 0.08$ \\
Population & $\mathbf{0.79 \pm 0.00}$ & $\mathbf{0.82 \pm 0.00}$ & $0.37 \pm 0.06$ & $0.36 \pm 0.11$ & $0.25 \pm 0.00$ & $0.15 \pm 0.00$ & $0.46 \pm 0.02$ & $0.79 \pm 0.00$ & $0.46 \pm 0.03$ \\
\midrule
\textbf{Classification} & \% Accuracy $\uparrow$ & & & & & & & & \\
\cmidrule(lr){1-2}
Countries & $94.28 \pm 0.18$ & $\mathbf{96.00 \pm 0.14}$ & $77.78 \pm 1.66$ & $82.11 \pm 1.72$ & $70.35 \pm 0.06$ & $67.80 \pm 0.03$ & $76.16 \pm 0.50$ & $90.72 \pm 0.44$ & $82.94 \pm 2.23$ \\
iNaturalist & $\mathbf{65.69 \pm 0.18}$ & $66.22 \pm 0.40$ & $56.73 \pm 0.83$ & $60.47 \pm 0.56$ & $58.78 \pm 0.48$ & $53.39 \pm 0.67$ & $56.73 \pm 0.8$ & $62.01 \pm 0.59$ & $60.83 \pm 0.53$ \\
Biome & $92.23 \pm 0.26$ & $\mathbf{94.41 \pm 0.14}$ & $75.81 \pm 1.53$ & $73.18 \pm 5.58$ & $69.69 \pm 0.06$ & $68.29 \pm 0.11$ & $79.61 \pm 0.42$ & $89.57 \pm 0.45$ & $83.55 \pm 2.43$ \\
Ecoregions & $89.32 \pm 0.31$ & $\mathbf{91.67 \pm 0.15}$ & $76.87 \pm 1.27$ & $78.43 \pm 1.71$ & $68.46 \pm 0.06$ & $67.26 \pm 0.02$ & $70.48 \pm 0.21$ & $84.65 \pm 0.32$ & $77.07 \pm 2.54$ \\

\bottomrule
\end{tabular}}
\end{table*}

\cref{fig:res:full} shows results (signed errors) for air temperature, population density and country code prediction, where the improvements from SatCLIP are visually apparent. \cref{tab:res_all} shows test set results on all datasets. One interesting observation is that this use of SatCLIP embeddings are more informative for iNat classification than a location encoder pretrained on iNat (CSP-iNat). This is intuitive as SatCLIP embeddings might be able to provide auxiliary information not contained within the iNat imagery. Overall, the results confirm that SatCLIP models trained on SK-100K data provide meaningful features to help with prediction in both natural (e.g., Air Temp., Elevation) and socio-economic (e.g., Med. Income, Cali. Housing) settings.

\subsection{Additional Results: Geographic Generalization}

\begin{table*}[t]%[ht]
\centering
\caption{\textbf{Geographic adaptation capabilities of SatCLIP (with ViT16 vision encoder) vs.\ baseline location embeddings to new geographic areas with no ($*$) or very few ($\dagger$) samples  from the held-out test continent.}  We report average test set $R^2$ and accuracy in \% $\pm 1$ standard deviation across $10$ independently initialized fine-tuning runs.}
\label{tab:res_adaptation_all}
\footnotesize
\resizebox{\textwidth}{!}{
\begin{tabular}{lrrrrrrrrrrr}
\toprule
 & \textbf{SatCLIP$_{L=10}$} & \textbf{SatCLIP$_{L=40}$} & \textbf{CSP} & \textbf{CSP} & \textbf{GPS2Vec} & \textbf{GPS2Vec} & \textbf{MOSAIKS} & \textbf{GeoCLIP} & \textbf{Identity} \\
\textbf{Test Continent} &  (S2-100K) &  (S2-100K) & (FMoW) & (iNat) & (tag) & (visual) & (Planet) & (MP-16) & ($y \sim g(\mathbf{c})$) \\
\midrule
\textbf{Asia}     &  &  &            &            &               &                               &            & & \\
\midrule
Air Temp.$^*$ R$^2$ $\mathbf{\uparrow}$ & $\mathbf{0.75 \pm 0.05}$ & $0.63 \pm 0.04$ & $0.09 \pm 0.37$ & $-0.50 \pm 1.32$ & $-3.95 \pm 4.89$ & $-0.23 \pm 0.17$ & $-2.13 \pm 3.50$ & $\mathbf{0.77 \pm 0.28}$ & $0.20 \pm 1.64$ \\
Elevation$^*$  & $\mathbf{0.46 \pm 0.09}$ & $\mathbf{0.48 \pm 0.07}$ & $-0.20 \pm 0.07$ & $-0.26 \pm 0.03$ & $-0.29 \pm 0.01$  & $-0.25 \pm 0.00$ & $-0.07 \pm 0.06$ & $\mathbf{0.50 \pm 0.03}$ & $-0.16 \pm 0.06$ \\
Pop. Density$^*$ & $\mathbf{0.42 \pm 0.08}$ & $\mathbf{0.45 \pm 0.04}$ & $-0.29 \pm 0.11$ & $-1.02 \pm 0.32$ & $-0.37 \pm 0.04$  & $-0.57 \pm 0.01$ & $0.05 \pm 0.12$ & $0.38 \pm 0.04$ & $0.03 \pm 0.07$ \\
\cmidrule(lr){1-1}
Countries$^\dagger$ \% Acc. $\uparrow$  & $\mathbf{36.90 \pm 4.32}$ & $19.17 \pm 2.82$ &  $1.22 \pm 0.05$ &  $1.28 \pm 0.01$ &  $1.12 \pm 0.00$ &  $0.92 \pm 0.02$ &  $1.56 \pm 0.47$ & $23.12 \pm 2.50$ & $1.24 \pm 0.12$ \\
iNaturalist$^*$  & $19.60 \pm 0.78$ & $\mathbf{20.91 \pm 0.77}$ & $19.85 \pm 0.55$ & $\mathbf{21.49 \pm 0.85}$ & $17.52 \pm 0.38$ & $18.11 \pm 0.34$ & $16.14 \pm 0.42$ & $\mathbf{20.94 \pm 0.38}$ & $\mathbf{21.08 \pm 0.69}$ \\
Biome$^*$      & $25.89 \pm 2.79$ & $16.44 \pm 1.21$ &  $1.98 \pm 0.62$ &  $3.00 \pm 2.60$ &  $1.76 \pm 0.04$ &  $2.79 \pm 0.19$ & $\mathbf{37.81 \pm 4.47}$ & $31.67 \pm 1.91$ & $6.24 \pm 2.71$ \\
Ecoregions$^\dagger$  & $\mathbf{21.02 \pm 1.09}$ & $10.86 \pm 1.19$ &  $1.55 \pm 0.17$ &  $1.41 \pm 0.14$ &  $1.49 \pm 0.03$ &  $1.48 \pm 0.00$ &  $1.36 \pm 0.10$ & $6.65 \pm 1.03$ & $1.52 \pm 0.47$ \\

\midrule
\textbf{Africa}     &  &  &            &            &               &                               &            & & \\
\midrule
Air Temp.$^*$ R$^2$ $\uparrow$ & $-4.71 \pm 2.29$ & $\mathbf{-1.48 \pm 0.70}$ & $-3.64 \pm 3.18$ & $-2.67 \pm 5.80$ & $-7.91 \pm 0.04$ & $-7.29 \pm 0.27$ & $-17.43 \pm 18.37$ & $-9.91 \pm 28.82$ & $-27.36 \pm 39.46$ \\
Elevation$^*$ & $-1.80 \pm 1.74$ & $\mathbf{-0.21 \pm 0.09}$ & $-0.39 \pm 0.09$ & $-1.20 \pm 0.55$ & $-0.13 \pm 0.06$ & $-0.15 \pm 0.02$ &  $-0.79 \pm 0.43$ & $-0.34 \pm 0.10$ & $-2.43 \pm 2.67$ \\
Pop. Density$^*$ & $0.17 \pm 0.12$ & $0.18 \pm 0.09$ & $-0.53 \pm 0.21$ & $-0.31 \pm 0.16$ & $-0.34 \pm 0.02$ & $-0.37 \pm 0.01$ &  $0.15 \pm 0.05$ & $\mathbf{0.32 \pm 0.03}$ & $-0.50 \pm 0.34$ \\
\cmidrule(lr){1-1}
Countries$^\dagger$ \% Acc. $\uparrow$ & $\mathbf{30.65 \pm 4.23}$ & $10.22 \pm 1.62$ & $0.47 \pm 0.01$ & $0.45 \pm 0.04$ & $0.47 \pm 0.01$ & $0.45 \pm 0.00$ & $0.48 \pm 0.00$ & $10.32 \pm 2.75$ & $2.74 \pm 2.52$ \\
iNaturalist$^*$ & $\mathbf{9.53 \pm 0.57}$ & $6.23 \pm 0.47$ & $6.63 \pm 0.57$ & $8.65 \pm 0.52$ & $7.47 \pm 0.53$ & $6.85 \pm 0.39$ & $5.18  \pm 0.38$ & $7.69 \pm 0.30$ & $\mathbf{9.96 \pm 0.33}$ \\
Biome$^*$  & $35.72 \pm 5.48$ & $12.34 \pm 1.75$ & $0.94 \pm 0.00$ & $1.09 \pm 0.48$ & $1.29 \pm 0.04$ & $1.17 \pm 0.21$ & $\mathbf{49.86 \pm 1.57}$ & $28.28 \pm 3.06$ & $1.46 \pm 0.67$ \\
Ecoregions$^\dagger$ & $\mathbf{32.03 \pm 1.19}$ & $12.91 \pm 1.63$ & $0.90 \pm 0.00$ & $0.94 \pm 0.04$ & $0.88 \pm 0.01$ & $0.90 \pm 0.00$ & $0.92  \pm 0.12$ & $12.41 \pm 2.20$ & $7.72 \pm 3.93$ \\
\midrule
\# of wins & 8 & 5 & 0 & 1 & 0 & 0 & 2 & 4 & 2 \\
\bottomrule
\end{tabular}}
\end{table*}

\begin{figure*}[ht]
    \centering
    \includegraphics[scale=0.5]{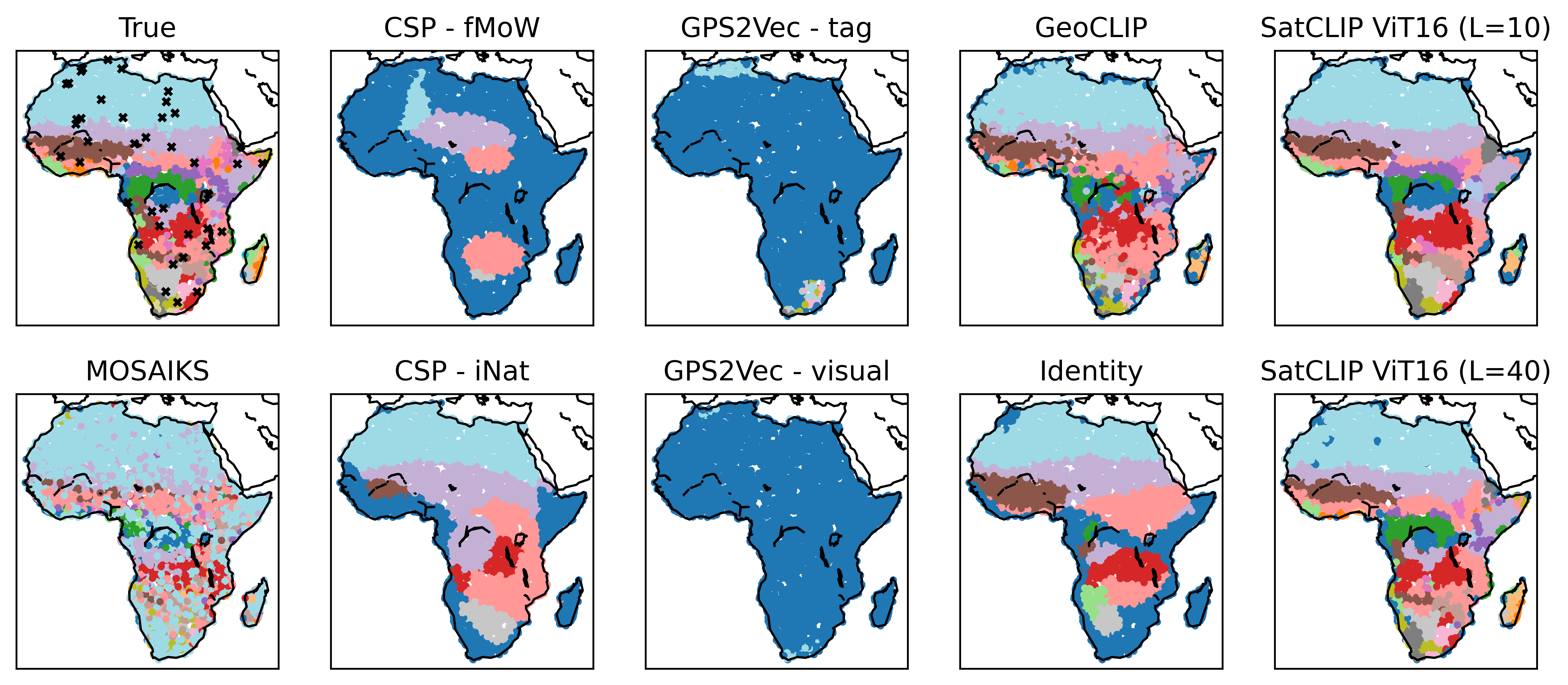}
    \caption{\textbf{Few-shot ecoregion prediction.} Predictions on the test continent Africa are shown for the different embeddings. Models are trained on only $1\%$ (marked by black crosses) of the data points on the test continent to evaluate their capacity for generalizing to an unseen environment.}
    \label{fig:res_continents_ecoregions}
\end{figure*}

\cref{fig:res_continents_ecoregions} shows results for the few-shot domain generalization setting with the Ecoregions dataset. Here, we highlight true values and predictions on the test set continent Africa. We can see that SatCLIP embeddings perform best, followed by GeoCLIP and MOSAIKS embeddings. This experiment helps us to evaluate the embeddings capacity for overcoming geographic distribution shift. The full results table for this experiment can be found in \cref{tab:res_adaptation_all}.

\subsection{Additional Results: Combining Embeddings}

We test whether combinations of embeddings obtained from location encoders trained on different datasets further improves performance. We do this by concatenating the embeddings before feeding them to the downstream learner. We show in our results in \cref{tab:res_combined} that combinations do not help and performance is not improved over the respective best single embedding.

\begin{table*}[ht]
\centering
\caption{\textbf{Downstream task performance using combinations of different embeddings.} We report average test set MSE ($\downarrow$) $\pm 1$ standard deviation across $10$ independently initialized MLP training runs on all regression tasks. Here, SatCLIP corresponds to a model with ResNet-50 vision encoder.}
\label{tab:res_combined}
\begin{adjustbox}{width=\textwidth}
\begin{tabular}{ll|lllll}
\toprule
& & Task & & & & \\
\textbf{Embedding $1$} & \textbf{Embedding $2$} & \textbf{Air Temp.} & \textbf{Med. Income} & \textbf{Cali. Housing} & \textbf{Elevation} & \textbf{Population} \\
\midrule
SatCLIP$_L=10$ & GeoCLIP & $0.23 \pm (0.01)$ & $0.67 \pm (0.01)$ & $1.81 \pm (0.11)$ & $0.14 \pm (0.01)$ & $0.48 \pm (0.01)$ \\
SatCLIP$_L=10$ & CSP (iNat) & $0.23 \pm (0.02)$ & $1.08 \pm (0.03)$ & $4.04 \pm (0.06)$ & $0.26 \pm (0.01)$ & $0.69 \pm (0.02)$ \\
SatCLIP$_L=10$ & GPS2Vec (tag) & $0.26 \pm (0.02)$ & $0.95 \pm (0.02)$ & $3.76 \pm (0.06)$ & $0.22 \pm (0.01)$ & $0.6 \pm (0.01)$ \\
SatCLIP$_L=10$ & GPS2Vec (visual) & $0.31 \pm (0.02)$ & $0.97 \pm (0.02)$ & $3.82 \pm (0.1)$ & $0.24 \pm (0.02)$ & $0.61 \pm (0.01)$ \\
CSP (iNat) & GPS2Vec (tag) & $1.56 \pm (0.08)$ & $1.29 \pm (0.02)$ & $5.66 \pm (0.09)$ & $0.88 \pm (0.04)$ & $2.06 \pm (0.16)$ \\
CSP (iNat) & GPS2Vec (visual) & $1.85 \pm (0.22)$ & $1.3 \pm (0.01)$ & $5.72 \pm (0.07)$ & $0.91 \pm (0.02)$ & $1.78 \pm (0.04)$ \\
GPS2Vec (visual) & GPS2Vec (tag) & $2.99 \pm (0.18)$ & $1.19 \pm (0.02)$ & $2.53 \pm (0.13)$ & $0.9 \pm (0.0)$ & $2.03 \pm (0.01)$ \\
\bottomrule
\end{tabular}
\end{adjustbox}
\end{table*}

\end{document}